\documentclass[10pt,twocolumn,letterpaper]{article}

\usepackage[pagenumbers]{iccv} % To force page numbers, e.g. for an arXiv version

\usepackage{graphicx}
\usepackage{amsmath}
\usepackage{amssymb}
\usepackage{mathtools} % Better than amsmath

\usepackage{xcolor}
\usepackage{enumitem}
\usepackage{xspace}
\usepackage{booktabs}
\usepackage{colortbl}
\usepackage{multirow}
\usepackage{makecell}
\usepackage{lipsum} % Just for generating dummy text, can be removed
\usepackage{gensymb} % for \degree

\newcommand{\Tref}[1]{Table~\ref{#1}}

\newcommand{\fref}[1]{Fig.~\ref{#1}}
\newcommand{\Fref}[1]{Figure~\ref{#1}}

\usepackage[labelsep=period]{caption}
\renewcommand{\paragraph}[1]{\vspace{0.2em}\noindent \textbf{#1 \hspace{0.2em}}}

\definecolor{MyDarkRed}{rgb}{0.46, 0.16, 0.16}
\definecolor{MyDarkBlue}{rgb}{0.16, 0.16, 0.66}

\newcommand{\suppl}{supplementary materials\xspace}

\newcommand{\MethodName}{LHM\xspace}
\newcommand{\HeadModule}{HFPE\xspace}

\newcommand{\headmodulefull}{head feature pyramid encoding\xspace}
\newcommand{\MMTBlock}{Multimodal Body-Head Transformer Block\xspace}

\newcommand{\MMTBlockabs}{MBHT-block\xspace}

\newcommand{\Frst}[1]{\textcolor{red}{\textbf{#1}}}
\newcommand{\Scnd}[1]{\textcolor{blue}{\textbf{#1}}}

\definecolor{iccvblue}{rgb}{0.21,0.49,0.74}
\usepackage[pagebackref,breaklinks,colorlinks,allcolors=iccvblue]{hyperref}

\newcommand{\wlink}[1]{\textcolor{magenta}{{#1}}}

\def\paperID{2249} % *** Enter the Paper ID here
\def\confName{ICCV}
\def\confYear{2025}

\title{LHM: Large Animatable Human Reconstruction Model \\
for Single Image to 3D in Seconds}
\author{Lingteng Qiu$\footnotemark[1]$ \quad Xiaodong Gu$\footnotemark[1]$ \quad Peihao Li$\footnotemark[1]$ \quad Qi Zuo$\footnotemark[1]$ \\ Weichao Shen \quad Junfei Zhang \quad Kejie Qiu \quad Weihao Yuan \quad \\  Guanying Chen$\footnotemark[2]$ \quad Zilong Dong $\footnotemark[2]$ \quad Liefeng Bo
\vspace{0.3em} \\
{Tongyi Lab, Alibaba Group} \\
}

\begin{document}

\twocolumn[{
\renewcommand\twocolumn[1][]{#1}
\maketitle
\vspace{-15pt}
\begin{center}
    \captionsetup{type=figure}
    \includegraphics[width=\textwidth]{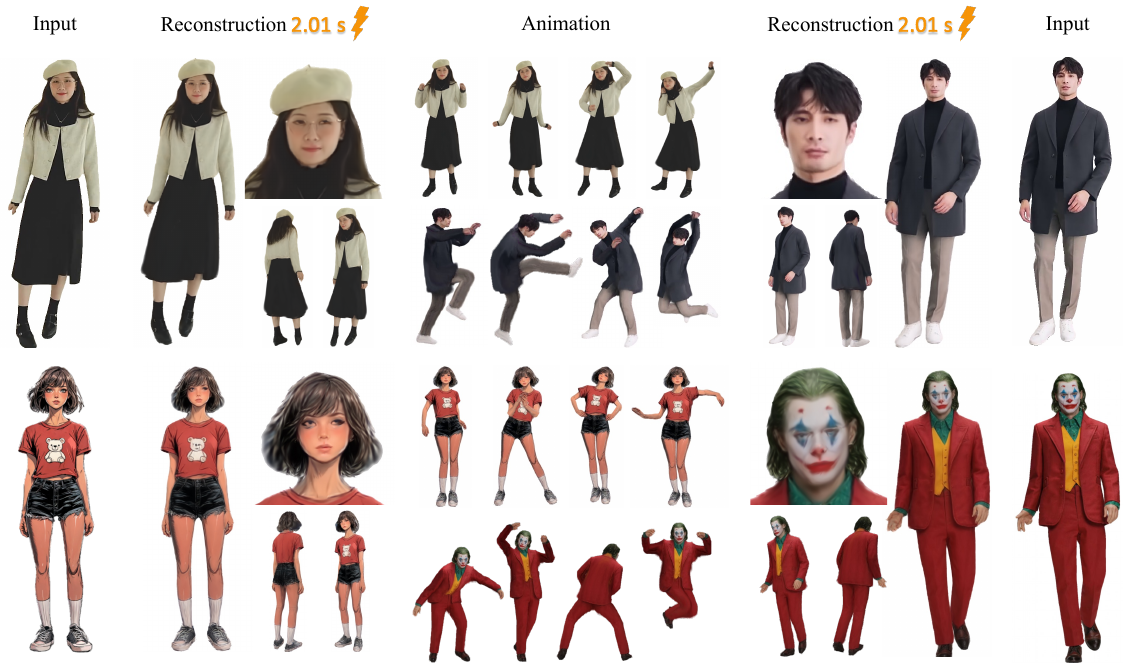}
    \\
    \captionof{figure}{\textbf{3D Avatar Reconstruction and Animation Results of our \emph{\MethodName}}. 
    Our method reconstructs an animatable human avatar in a single feed-forward pass in seconds.
    The resulting model supports real-time rendering and pose-controlled animation.
    }
    \label{fig:teaser}
\end{center}
%\maketitle
}]

\footnotetext[1]{Equal contribution.}
\footnotetext[2]{Corresponding author.}

%%%%%%%%% ABSTRACT
\begin{abstract}

Animatable 3D human reconstruction from a single image is a challenging problem due to the ambiguity in decoupling geometry, appearance, and deformation. 
Recent advances in 3D human reconstruction mainly focus on static human modeling, and the reliance of using synthetic 3D scans for training limits their generalization ability.
Conversely, optimization-based video methods achieve higher fidelity but demand controlled capture conditions and computationally intensive refinement processes. 
Motivated by the emergence of large reconstruction models for efficient static reconstruction, we propose LHM (Large Animatable Human Reconstruction Model) to infer high-fidelity avatars represented as 3D Gaussian splatting in a feed-forward pass.
Our model leverages a multimodal transformer architecture to effectively encode the human body positional features and image features with attention mechanism, enabling detailed preservation of clothing geometry and texture.
To further boost the face identity preservation and fine detail recovery, we propose a  head feature pyramid encoding scheme to aggregate multi-scale features of the head regions.
Extensive experiments demonstrate that our LHM generates plausible animatable human in seconds without post-processing for face and hands, outperforming existing methods in both reconstruction accuracy and generalization ability. Our code is available on 
\href{https://github.com/aigc3d/LHM}{\wlink{https://github.com/aigc3d/LHM}
}.
\end{abstract}

%Existing approaches typically rely on parametric human models, which offer strong structural priors but fail to capture fine-grained clothing details, 
%    or optimization-based methods, which require extensive computational time to refine geometry and appearance. \todo{What are the main problems of existing methods}

\section{Introduction}
\label{sec:intro}
Creating 3D animatable human avatars from single images is crucial for immersive AR/VR applications, yet remains challenging due to the coupled ambiguities of geometry, appearance, and deformation.

Recently, diffusion-based human video animation methods~\cite{hu2024animate, zhu2024champ, lin2025omnihuman, men2024mimo} have shown the capability to generate photorealistic human videos. However, these methods often suffer from inconsistent views under extreme poses and require long inference times for video sampling.
In 3D animatable human reconstruction,
early methods rely on parametric models~\cite{loper2015smpl, alldieck2018videobasedreconstruction3d} to provide strong human body priors for animation but struggle to capture the fine-grained geometry of loose clothing, and high-fidelity facial details, limiting their expressiveness.
%\weihao{only this one limitation? list more. e.g., high resolution texture especially on face} 
While learning-based 3D methods have made considerable progress in static clothed human reconstruction in recent years~\cite{saito2019pifu,saito2020pifuhd,xiu2023econexplicitclothedhumans}, most existing approaches either fail to produce animatable humans or lack generalization to in-the-wild images~\cite{huang2020archanimatablereconstructionclothed,he2022archanimationreadyclothedhuman}.
%\weihao{need to mark that here is for 3D methods. parametric model-based methods also include some 2D methods.}
The problem of generalizable 3D animatable human reconstruction from a single image remains underexplored due to the lack of a suitable model architecture and a large-scale 3D rigged human dataset for learning.
%\weihao{suggestion: first talk about 2D methods (good results and weakness), then talk about 3D methods (advantages and disadvantages), otherwise 3D-2D-3D is confusing}

Recently, large reconstruction model (LRM)~\cite{hong2023lrm} have shown that scalable transformer models trained with large-scale of 3D data can learn generalizable priors for single-image 3D object reconstruction. 
While promising, extending this success to \emph{animatable human reconstruction} presents unique challenges that demand solutions to two critical problems: 1) designing an effective architecture that combines 3D human representation with animation capabilities, and 2) overcoming the scarcity of high-quality 3D rigged human training data.

In this work, we propose \emph{LHM} (Large Animatable Human Model), a scalable feed-forward transformer model, that produces animatable 3D human avatars in seconds from single images. 
We represent the human avatar as Gaussian splatting considering its real-time photo-realistic rendering. 
Our method takes a single image as input and directly predicts the canonical human as a set of 3D Gaussians in canonical space.
To enable animation, our method starts from a set of surface points sampled from the SMPL-X~\cite{smplx:2019} template mesh to serve as geometric anchors for predicting 3D Gaussian properties.

The network architecture of our method is inspired by the multimodal transformer (MM-transformer) block introduced by the state-of-the-art image generation model SD3~\cite{esser2024scaling}, which is designed to model the interaction between the text and image tokens. 
We adapt the MM-transformer to our task to effectively encode the body 3D point features and 2D image features with attention operation, enabling detailed preservation of clothing geometry and texture.
To address the problem of facial identity preservation, we further introduce a \emph{\headmodulefull} (\HeadModule) scheme that aggregates multi-scale visual features from head regions, significantly improving detail recovery in these perceptually sensitive areas.

During training, to boost the performance of this transformer from large-scale data, we transform the predicted canonical Gaussians to various poses using SMPL-X skeleton parameters and optimize through a combination of rendering losses and regularizations. This self-supervised strategy allows learning generalizable human priors from readily available video data rather than scarce 3D scans.

In summary, our contributions are:
\begin{itemize}[itemsep=0pt,parsep=0pt,topsep=2bp]
    \item We introduce a generalizable large human reconstruction model that produces animatable 3D avatars from single images in seconds.
    \item We propose a multimodal human transformer to fuse 3D surface point features and image features via an attention mechanism, enabling joint reasoning across geometric and visual domains.
    \item Our method, trained on a large-scale video dataset without rigged 3D data, achieves state-of-the-art performance on real-world imagery, surpassing existing approaches in generalization and animation consistency.
\end{itemize}

\section{Related Work}
\label{sec:related_work }

\subsection{Single-Image Human Reconstruction}
Early approaches to single-image 3D human reconstruction primarily relied on parametric body models like SMPL~\cite{loper2015smpl,smplx:2019} to predict geometric offsets for naked or clothed subjects~\cite{choutas2022accurate3dbodyshape,kanazawa2018endtoendrecoveryhumanshape, alldieck2018detailedhumanavatarsmonocular,alldieck2019tex2shapedetailedhumanbody}. 
These methods often struggle to capture diverse clothing styles due to their rigid mesh-based representations. 
Subsequent advancements leveraged implicit functions~\cite{saito2019pifu,saito2020pifuhd,xiu2023econexplicitclothedhumans,cao2022jiff,zheng2020pamirparametricmodelconditionedimplicit,zhang2024sifusideviewconditionedimplicit, xiong2024mvhumannet, bib:havefun} to model complex geometries, providing greater flexibility in handling fine surface details without resolution constraints. 
Generative frameworks, such as diffusion models~\cite{chen2024ultramansingleimage3d,he2024magicmangenerativenovelview, Pang2024Disco4DD4} and GANs~\cite{men2024en3denhancedgenerativemodel, kolotouros2024dreamhuman}, have been adopted to synthesize detailed 3D humans conditioned on the input images. Some researchers~\cite{huang2024tech,xiu2024puzzleavatar,XAGen2023,huang2023humannorm} have explored generating avatars distilled from diffusion models using SDS loss. However, these methods tend to be time-consuming.

\begin{figure*}[tb] \centering
    \includegraphics[width=0.95\textwidth]{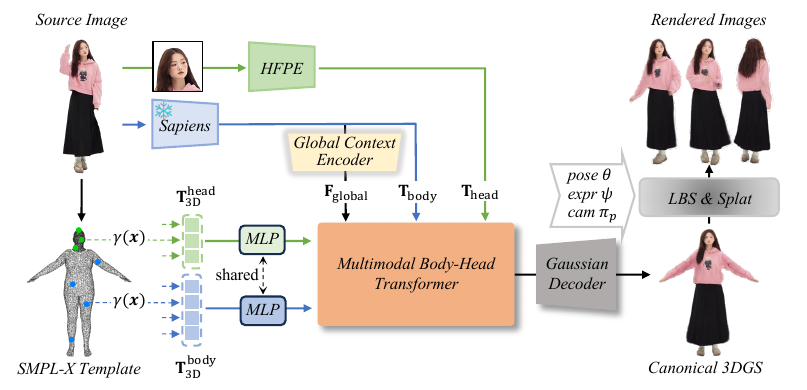}
    \caption{\textbf{Overview of the proposed \emph{\MethodName}.}
    Our method extracts body and head image tokens from the input image, and utilizes the proposed Multimodal Body-Head Transformer (MBHT) to fuse the 3D geometric body tokens with the image tokens. After the attention-based fusion process, the geometric body tokens are decoded into Gaussian parameters.
    } \label{fig:pipeline}
\end{figure*}

Recently, large reconstruction models (LRMs) have been introduced to enable generalizable feed-forward object reconstruction~\cite{hong2023lrm,tang2025lgm}, significantly accelerating inference time. 
Human-LRM~\cite{weng2024template} employs a feed-forward transformer to decode triplane-NeRF representations to enable multi-view rendering, followed by diffusion-based 3D reconstruction. 
Human-Splat~\cite{pan2025humansplat} first generates multi-view images using video diffusion, and then applies a latent reconstruction transformer to predict human avatars in 3D Gaussian Splatting (3DGS). PSHuman~\cite{li2024pshuman} utilizes multi-view diffusion with the ID diffusion to improve the face quality.
In contrast to these methods, which focus on static human reconstruction, our approach generates photorealistic, animatable human avatars, enabling high-quality dynamic rendering and interaction.

\subsection{Animatable Human Generation}
Creating animatable avatars has evolved from parametric-model-driven methods~\cite{alldieck2018videobasedreconstruction3d} to hybrid approaches combining implicit surfaces and human priors for clothed body modeling~\cite{he2022archanimationreadyclothedhuman}. 
Video-based techniques further improved reconstruction consistency by leveraging temporal cues from monocular~\cite{jiang2022selfrecon,weng2022humannerf,qiu2023recmv, hu2023gauhuman, tan2025dressrecon, yu2023monohuman} or multi-view sequences~\cite{chen2024meshavatar,li2024animatable,peng2021animatable, moon2024exavatar, li2024animatablegaussians}. 
The rise of text-to-3D methods~\cite{poole2022dreamfusion} has enabled avatar generation from text prompts through a long optimization process~\cite{huang2024dreamwaltz,kolotouros2024dreamhuman,cao2024dreamavatar}.

In 3D animatable human reconstruction, CharacterGen~\cite{peng2024charactergen} has explored cartoon-style avatar generation using diffusion models for canonical view synthesis and transformer-based shape reconstruction. 
AniGS~\cite{qiu2024AniGS} employs a video diffusion model to generate multi-view canonical images, followed by a 4D Gaussian splatting (4DGS) optimization to achieve consistent 3D generation. 
GAS~\cite{lu2025gas} adopts a generalizable human NeRF to reconstruct the subject in a canonical space, followed by video diffusion for refinement.
More recently, IDOL~\cite{zhuang2024idol} introduces a feed-forward transformer model to decode Gaussian attribute maps in UV spaces, and requires post-processing to refine the face and hands.
Unlike these methods, we propose an efficient and expressive multimodal human transformer that directly regresses 3D human Gaussians without relying on UV representations or requiring post-processing for hands and face refinement.

%\todo{double check the citation format in references}
%RIn contrast, our approach prioritizes real-time rendering and robustness by leveraging explicit 3D Gaussian representations conditioned on learned geometry distributions, circumventing the limitations of mean-seeking implicit models~\cite{liu2021neural, kwon2024deliffas}.

\section{Preliminary}
\paragraph{Human Parametric Model} 
\label{sec:preliminary-human-model}
The SMPL~\cite{loper2015smpl} and SMPL-X~\cite{smplx:2019} parametric models are commonly employed for representing human body structures in computer vision applications. These frameworks employ advanced deformation techniques, including skinning mechanisms and blend shapes, which are derived from extensive datasets comprising thousands of 3D body scans.

In particular, the SMPL-X model utilizes two primary types of parameters to capture human body configurations. These include shape parameters~\(\boldsymbol{\beta} \in \mathbb{R}^{20}\), and pose parameters~\(\boldsymbol{\theta} \in \mathbb{R}^{55 \times 3}\),  which determine the articulation and deformation of the body mesh based on skeletal poses.

\paragraph{3D Gaussian Splatting} The 3D Gaussian Splatting framework~\cite{kerbl3Dgaussians} models three-dimensional information through a set of anisotropic Gaussian distributions. Each primitive is parameterized by a centroid \(\mathbf{p} \in \mathbb{R}^3\), scale parameters \(\mathbf{\sigma} \in \mathbb{R}^3\), and a quaternion \(\mathbf{r} \in \mathbb{R}^4\) representing rotation. The model also incorporates an opacity parameter \(\rho \in [0,1]\) and appearance features \(\mathbf{f} \in \mathbb{R}^C\) encoded with spherical harmonics to account for view-dependent lighting effects. During differentiable rendering, these volumetric primitives are projected into 2D screen space as oriented Gaussian distributions, followed by view-ordered alpha blending to composite the final pixel colors. 

\section{Method}
\subsection{Overview}
\label{sub:Overview}

Given an input RGB human image \( I \in \mathbb{R}^{H \times W \times 3} \), our goal is to reconstruct an animatable 3D human avatar in seconds. The avatar is represented via 3D Gaussian Splatting (3DGS), which supports real-time, photorealistic rendering and pose-controlled animation. To achieve this, we propose \emph{Large Animatable Human Reconstruction Model (LHM)}, a feed-forward, transformer-based architecture that directly predicts the canonical 3D avatar from a single image.

Inspired by recent multimodal transformers~\cite{esser2024scaling}, we design a  \emph{Multimodal Body-Head Transformer} (MBHT) to effectively integrate geometric and image features.
As illustrated in \fref{fig:pipeline}, our framework processes training pairs that consist of a source image, a target view image, a foreground mask, SMPL-X pose parameters, and a camera matrix. 

The proposed MBHT employs attention operations to integrate three types of tokens: geometric tokens, body image tokens, and head image tokens, where geometric tokens can effectively attend to other tokens, allowing local and global refinement. In addition, Body and head tokens interact through a part-aware transformer, ensuring balanced attention across different body regions.

After the attention-based token fusion process, the geometric body tokens are decoded into per-Gaussian parameters, including deformation, scaling, rotation, and spherical harmonic (SH) coefficients. During training, we employ Linear Blend Skinning (LBS) to warp the canonical avatar to the target view, where photometric and regularization losses guide the learning process.

%As illustrated in \fref{fig:pipeline}, our framework processes training pairs that consist of a source image \( I_{\text{src}} \), a target view image \( I_{\text{tgt}} \), a foreground mask \( M_{\text{tgt}} \), SMPL-X pose parameters \( \theta_{\text{tgt}} \), and a projection matrix \( \pi_p \). 

%As shown in \fref{fig:pipeline}, our unified framework processes training pairs containing: a source image \( I_{\text{src}} \), target view image \( I_{\text{tgt}} \), foreground mask \( M_{\text{tgt}} \), SMPL-X pose parameters \( \theta_{\text{tgt}} \), and projection matrix \( \pi_p \). Our novel part-aware human transformer architecture (detailed in \sref{sec:architecture}) operates on these inputs along with visual tokens, positional encodings, and identity embeddings to predict 3DGS representations in canonical space.
%
%For end-to-end training, we employ Linear Blend Skinning (LBS) to transform the canonical avatar into target view space. The transformed Gaussian model is then rendered into the target viewpoint, where we apply: 1) a photometric loss comparing rendered and ground truth colors, and 2) carefully designed regularization terms to regularize 3DGS in canonical space~(details in \sref{sec:target_function}).

\subsection{Geometric and Image Feature Encoding}
\label{sec:architecture}

Geometric tokens are derived from SMPL-X surface points, encoding structural priors of the human body. Body image tokens are extracted from a pretrained vision transformer~\cite{khirodkar2024sapiens}, encoding texture and appearance. Head tokens specialize in capturing high-frequency facial details through a multi-scale feature extraction process.

\paragraph{Human Geometric Feature Encoding}
Leveraging SMPL-X's human body prior, we initialize 3D query points \( \{\mathbf{x}_i\}_{i=1}^{N_{\text{points}}} \subset \mathbb{R}^3 \) by strategically sampling the canonical pose mesh. Each point undergoes positional encoding~\cite{mildenhall2020nerf}, followed by a multi-layer perceptron (MLP) projection to match the token channel dimension of the transformer:
\begin{equation}
    \mathbf{T}_{\text{3D}} = \text{MLP}_{\text{proj}}\left(\gamma(\mathbf{X})\right) \in \mathbb{R}^{N_{\text{points}} \times C},
\end{equation}
where \( \gamma: \mathbb{R}^3 \to \mathbb{R}^{3L} \) applies \( L \)-frequency sinusoidal encoding to spatial coordinates, and \( C \) is the token dimension.
%, and \( \text{MLP}_{\text{proj}} \) aligns dimensions with transformer expectations.

\begin{figure}[tb] \centering
    \includegraphics[width=0.48\textwidth]{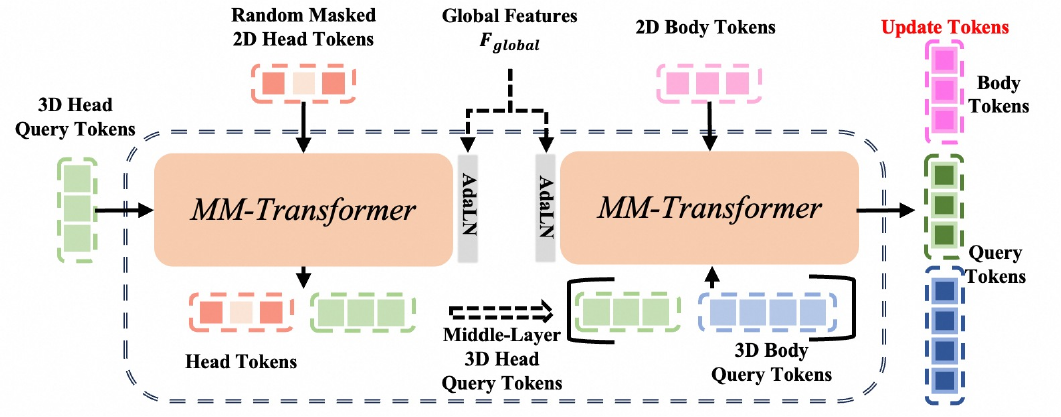}
    \caption{Architecture of the proposed \MMTBlock (MBHT-block).} 
    \label{fig:mm-transformer}
\end{figure}

\paragraph{Body Image Tokenization} 
Building on large-scale vision transformers pretrained on human-centric datasets~\cite{khirodkar2024sapiens}, we convert image pixels into transformer-compatible tokens. 
Specifically, we employ the frozen Sapiens-1B encoder $\mathcal{E}_{\text{Sapiens}}$, pretrained on 10 million human images, to extract semantic body features:
\begin{equation}
    \mathbf{T}_{\text{body}} = \text{MLP}_{\text{proj}}\left({\mathcal{E}_{\text{Sapiens}}(I)}\right) \in \mathbb{R}^{N_{\text{body}} \times C},
\end{equation}
where \( N_{\text{body}} \) denotes the body token numbers.

\paragraph{Head Feature Pyramid Tokenization}
However, since the human head occupies only a small region in the input image and undergoes spatial downsampling in the encoder, critical facial details are often lost. To mitigate this issue, we propose a \headmodulefull (\HeadModule) that aggregates multi-scale features \{$\mathcal{E}_\text{dino}^\cdot$\} from DINOv2~\cite{oquab2023dinov2}:
\begin{equation}
    \mathbf{T}_{\text{head}} = \mathcal{F}_{\text{fusion}}\left( \mathcal{E}_{\text{dino}}^4, \mathcal{E}_{\text{dino}}^{11}, \mathcal{E}_{\text{dino}}^{17}, \mathcal{E}_{\text{dino}}^{23} \right) \in \mathbb{R}^{N_{\text{head}} \times C},
\end{equation}
where \( \mathcal{F}_{\text{fusion}} \) fuses features from the 4th, 11th, 17th, and 23rd transformer blocks using depthwise concatenation and 1×1 convolutions, followed by feature projection. This design captures a hierarchy of semantic abstractions: early blocks retain high-frequency texture details, while deeper layers encode robust head geometry priors.
 %(see \cref{fig:pyramid_head_encoder})

\subsection{Multimodal Body-Head Transformer}
\paragraph{Global Context Feature}
The global context token is used for the modulation of the attention block.
To capture global context information for attention modulation, we take body tokens as the input, followed by max pooling and two MLP layers to extract the global context embeddings:
\begin{equation}
    \mathbf{F}_{\text{global}} = \text{MLP}_{\text{global}}\left(\text{MaxPool}\left(\mathbf{T}_{\text{body}} \right)\right).
\end{equation}

\paragraph{Multimodal Body-Head Transformer Block} 
The core design of the proposed model architecture is the multimodal body-head transformer block  (MBHT-block) that efficiently fuse 3D geometric tokens with the body and head image features, as shown in \fref{fig:mm-transformer}.

Specifically, the global context features, image tokens, and query point tokens are fed simultaneously into the MBHT-block. To enhance the learning of features specific to the head and body, the 3D head point tokens will first fuse with the head image features, and then concatenate with the 3D body point token to interact with the body image tokens. 
\begin{equation}
\begin{aligned}
    \mathbf{T}^\text{head}_{\text{3D}},\mathbf{T}_{\text{head}} & \coloneqq \text{MM-T}\left(\mathbf{T}^\text{head}_{\text{3D}}, \mathbf{T}_{\text{head}}; \mathbf{F}_{\text{global}}\right) \\
    \mathbf{T}_{\text{3D}} & \coloneqq \text{Norm}(\mathbf{T}^{\text{head}}_{\text{3D}}) ~\Vert~ \text{Norm}(\mathbf{T}^{\text{body}}_{\text{3D}}) \\
    \mathbf{T}_{\text{3D}},\mathbf{T}_{\text{body}} & \coloneqq \text{MM-T}\left(\mathbf{T}_{\text{3D}}, \mathbf{T}_{\text{body}}; \mathbf{F}_{\text{global}}\right)
\end{aligned}
\end{equation}
%\begin{equation}
%\begin{aligned}
%\mathbf{\tilde{T}}_{\text{phead}},\mathbf{T}_{\text{head}} &= \text{MM-T}\left(\mathbf{M}_{\text{head}} \odot \mathbf{T}_{\text{points}}, \mathbf{T}_{\text{head}}; \mathbf{F}_{\text{identity}}\right) \\
%    \mathbf{T}_{\text{points}} &= \text{Norm}(\mathbf{\tilde{T}}_{\text{phead}}) \Vert \text{Norm}((\mathbf{1}-\mathbf{M}_{\text{head}}) \odot \mathbf{T}_{\text{points}}) \\
%    \mathbf{T}_{\text{points}},\mathbf{T}_{\text{body}} &= \text{MM-T}\left(\mathbf{T}_{\text{points}}, \mathbf{T}_{\text{body}}; \mathbf{F}_{\text{identity}}\right)
%\end{aligned}
%\end{equation}
where $\mathbf{T}^\text{body}_{\text{3D}}$ and $\mathbf{T}^\text{head}_{\text{3D}}$ are 3D body and head points in $\mathbf{T}_{\text{3D}}$, respectively. MM-T indicates the Multimodal Transformer Block~\cite{esser2024scaling}, and \( \Vert \) denotes token-wise concatenation~(see \suppl for details).

\begin{figure*}[tb]
    \centering
    \includegraphics[width=1.0\textwidth]{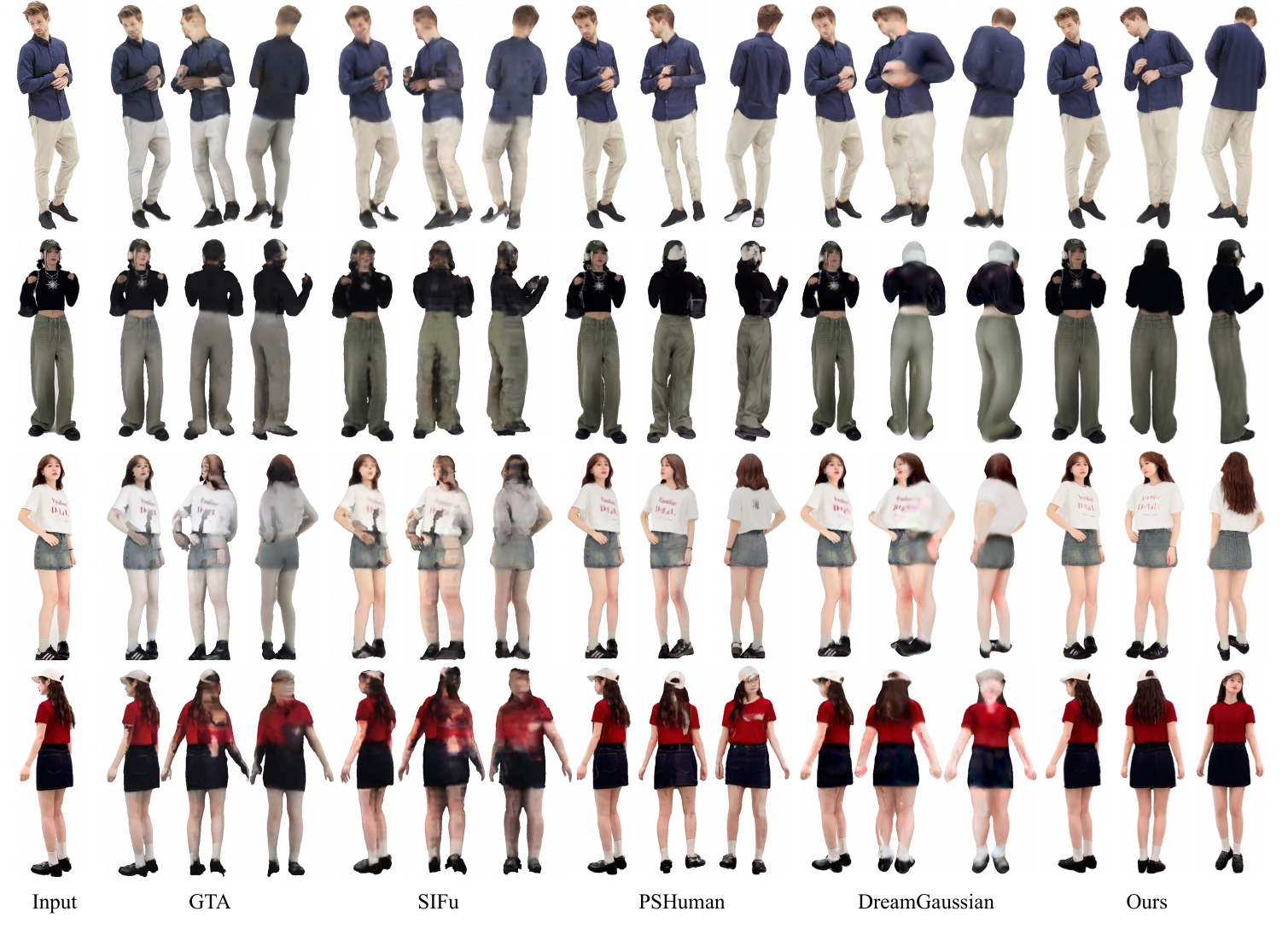}
    \\
    \vspace{-1em}
    \caption{
        Single-view reconstruction comparisons on DeepFashion~\cite{liuLQWTcvpr16DeepFashion} and in-the-wild images. 
        \MethodName achieves superior appearance fidelity and texture sharpness, particularly evident in facial details and garment wrinkles.
    }
    \label{fig:qualitative_on_static}
\end{figure*}

\paragraph{Head Token Shrinkage Regularization}  
Our experiments show that the attention mechanism in \MMTBlockabs relies heavily on head-region features, which limits its ability to learn body-part features effectively. To address this imbalance, we take inspiration from MAE~\cite{MaskedAutoencoders2022} and randomly mask the head region of the input crop during training.

Specifically, we apply spatial masking to head tokens with a ratio ranging from 0\% to 50\%, encouraging the model to compensate through enhanced body-context utilization. This regularization strategy improves body-part self-attention capacity while maintaining head reconstruction fidelity.

%In the experiments, we find the \MMTBlockabs's attention mechanism exhibits disproportionate reliance on head-region features, leading to diminished expressiveness in body-part feature learning. To mitigate this modality imbalance, inspired by MAE~\cite{MaskedAutoencoders2022},  we randomly mask the crop input head image area during training.
%Specifically, we apply range from 0\% to 50\% spatial masking to head tokens, forcing the model to compensate through enhanced body-context utilization. This regularization strategy improves body-part self-attention capacity while maintaining head reconstruction fidelity.

\paragraph{3DGS Parameter Prediction}
After \( N_{\text{layer}} \) \MMTBlockabs blocks, an MLP heads predict 3DGS parameters:
%$\mathbf{\chi}(\mathbf{p},\mathbf{r},\mathbf{f},\rho,\sigma)$:
\begin{equation}
\begin{aligned}
    \{\Delta\mathbf{p}_i, \mathbf{r}_i, \mathbf{f}_i, \rho_i, \sigma_i\} = \text{MLP}_{\text{regress}}(\mathbf{T}_{\text{points}}^{(i)}) \\
    \mathbf{p}_i = \mathbf{x}_i + \Delta\mathbf{p}_i, \quad \forall i \in \{1,...,N_{\text{points}}\}
\end{aligned}
\end{equation}
where \( \Delta\mathbf{p}_i \in \mathbb{R}^3 \) represents residual position offsets from the canonical SMPL-X.

%\begin{itemize}
%    \item denote vectors by upright bold letters: $\mathbf{l}, \mathbf{n}$
%    \item it's a Greek letter you have to use $\boldsymbol{l}, \boldsymbol{n}$ instead
%\end{itemize}

% \subsection{Loss Function}
% \label{sec:target_function}

% In this section, we formalize the end-to-end training objective function using only photometric supervision from in-the-wild video sequences and carefully designed regularization in canonical space.

% \paragraph{View Space Supervision} After obtaining the predicted avatar 3DGS parameters $\mathbf{\chi}(\mathbf{p},\mathbf{r},\mathbf{f},\rho,\sigma)$, we apply Linear Blend Skinning (LBS) to transform the avatar to view-space. Then, a predict image \predI and alpha mask \predM can be rendered at an target camera view $\pi_t$ through Gaussian splatting. To better model the dress deformation, we use diffused voxel skinning method~\cite{qiu2023recmv}~(details in \suppl).

% Finally, we employ L1 loss for both color $\mathcal{L}_\text{color}$ and mask supervision $\mathcal{L}_\text{mask}$. We also include a perceptual loss, \(\mathcal{L}_{\text{lpips}}\)  to promote the quality of rendering results:

% \begin{equation}
%     \mathbf{\mathcal{L}}_{\text{photometric}} = \lambda_{\text{rgb}} \mathcal{L}_\text{color} +  \lambda_{\text{mask}} \mathcal{L}_\text{mask} + \lambda_{\text{per}} \mathcal{L}_\text{lpips}
% \end{equation}
% in our setting, we set $\lambda_{\text{rgb}}$ is to 1, $\lambda_{\text{mask}}$ is to 0.5, and $\lambda_{\text{per}}$ is to 1. 

\subsection{Loss Function}
\label{sec:target_function}

Our training objective combines photometric supervision from in-the-wild video sequences with regularization constraints in canonical space. The complete optimization framework enables the learning of animatable avatars without requiring ground truth 3D supervision.

\subsubsection{View Space Supervision} 
Given the predicted 3DGS parameters $\mathbf{\chi} = (\mathbf{p}, \mathbf{r}, \mathbf{f}, \rho, \sigma)$, we first transform the canonical avatar to target view space using Linear Blend Skinning (LBS). The transformed Gaussian primitives are then rendered through differentiable splatting to produce the RGB image $\hat{I}$ and alpha mask $\hat{M}$ under target camera parameters $\pi_t$. To better model clothing deformations, we use a diffused voxel skinning~\cite{qiu2023recmv}.

The view-consistent supervision comprises three components in view space:
% \begin{equation}
%     \mathbf{\mathcal{L}}_{\text{photometric}} = \lambda_{\text{rgb}} \mathcal{L}_\text{color} +  \lambda_{\text{mask}} \mathcal{L}_\text{mask} + \lambda_{\text{per}} \mathcal{L}_\text{lpips}
% \end{equation}
\begin{equation}
    \mathcal{L}_{\text{photometric}} = \underbrace{\lambda_{\text{rgb}} \mathcal{L}_{\text{color}}}_{\text{Appearance}} + \underbrace{\lambda_{\text{mask}} \mathcal{L}_{\text{mask}}}_{\text{Silhouette}} + \underbrace{\lambda_{\text{per}} \mathcal{L}_{\text{lpips}}}_{\text{Perceptual Quality}}.
\end{equation}

In our implementation, the loss weights balance reconstruction aspects: $\lambda_{\text{rgb}} = 1.0$ for direct color supervison, $\lambda_{\text{mask}} = 0.5$ for geometric alignment, and $\lambda_{\text{per}} = 1.0$ to preserve high-frequency details.

% \paragraph{Regularization in Canonical Space} Although photometric loss provides strong supervision the final model in canonical loss, the model in canonical space is under-constraint, results in some artifacts when model is warp to novel pose. To address this problem,  we carefully design two regularization loss to constrain the Gaussian avatar model in canonical space.

% Firstly, to prevent degenerate Gaussian primitives with excessive anisotropy (e.g., needle-like ellipsoids), we use a as sphereical as possible~(ASAP) loss~\cite{physgaussian} $\mathcal{L}_{\text{asap}}$ to regularize their shape.

% Additionally, to prevent the mean position of Gaussian primitives leaving far away from initialized SMPL-X, we design a as close as possible~(ACAP) loss to regularize their position:

% \begin{equation}
% \begin{aligned}
%     \mathcal{L}_{\text{acap}}= \frac{1}{N_{\text{points}}}\sum_{i=0}^{N_{\text{points}}}\text{max}\{\Delta\mathbf{p}_i,d\}-d
% \end{aligned}
% \end{equation}
% where $d$ is experically threshold value to prevent the offset values large than the value.

% In summary, the regularization loss $\mathcal{L}_{\text{reg}}$ in canonical space is formulated as: 

% \begin{equation}
% \begin{aligned}
%     \mathcal{L}_{\text{reg}}= \lambda_{\text{asap}} \mathcal{L}_{\text{asap}} + \lambda_{\text{acap}} \mathcal{L}_{\text{acap}}
% \end{aligned}
% \end{equation}
% where $\lambda_{\text{asap}}$ is set to 10 while $\lambda_{\text{asap}}$ is set to 50.

\subsubsection{Canonical Space Regularization} 
While photometric loss provides effective supervision in target view space, the canonical representation remains under-constrained due to the ill-posed nature of monocular reconstruction. This limitation manifests as deformation artifacts when warping the avatar to novel poses. To address this fundamental challenge, we introduce two complementary regularization terms that enforce geometric coherence in canonical space.

\paragraph{Gaussian Shape Regularization} We apply the \emph{as spherical as possible loss} to penalize excessive anisotropy in Gaussian primitives:
\begin{equation}
    \mathcal{L}_{\text{ASAP}} = \frac{1}{N}\sum_{i=1}^N \| \mathbf{S}_i - \text{diag}(1) \|_F^2,
\end{equation}
where \(\mathbf{S}_i\) represents the covariance matrix, effectively discouraging needle-like ellipsoids while preserving necessary shape variation.

\paragraph{Positional Anchoring} To maintain body surface plausibility, we include the \emph{as close as possible loss} to encourage Gaussian positions to be close to their SMPL-X initialized locations by a hinged distance constraint:
\begin{equation}
    \mathcal{L}_{\text{ACAP}} = \frac{1}{N_{\text{points}}}\sum_{i=1}^{N_{\text{points}}} \max\left(\|\Delta\mathbf{p}_i\|_2 - d, 0\right),
\end{equation}
where \(d\) represents an empirically determined threshold (5.25cm) in practice) that permits local adjustments while preventing catastrophic drift.

The combined canonical regularization operates as:
\begin{equation}
    \mathcal{L}_{\text{reg}} = 50\mathcal{L}_{\text{ASAP}} + 10\mathcal{L}_{\text{ACAP}}.
    %\mathcal{L}_{\text{reg}} = \underbrace{50\mathcal{L}_{\text{ASAP}}}_{\text{Shape control}} + \underbrace{10\mathcal{L}_{\text{ACAP}}}_{\text{Positional stability}}.
\end{equation}
In summary, our composite training objective combines photometric fidelity preservation with geometric regularization, formulated as:
\begin{equation}
    \mathcal{L}_{\text{total}} = \mathcal{L}_{\text{photometric}} + \mathcal{L}_{\text{reg}}.
\end{equation}

\section{Experiments}
\label{sec:experiments}
\subsection{Implementation Details}
\paragraph{In-the-Wild Training Data} 
We curate a large-scale dataset of 301,733 single-person video sequences from 500K initial human motion footage samples collected from public video repositories. Our multi-stage filtering pipeline removes sequences containing multi-person interactions, occluded faces, or low-quality frames through manual inspection and automated metric thresholds.

\paragraph{Synthetic Data Augmentation} 
To address viewpoint bias in natural videos, we supplement the training with synthetic human scans from three sources: 2K2K~\cite{han2023highfidelity3dhumandigitization}, Human4DiT~\cite{shao2024human4dit} and RenderPeople~(see \suppl for details).

\paragraph{Preprocessing Pipeline} We employ SAMURAI~\cite{yang2024samurai} to extract foreground masks across video sequences. For SMPL-X parametric estimation, we leverage Multi-HMR~\cite{multi-hmr2024} to estimate pose and shape parameters.

\paragraph{Training Configuration}
Our implementation utilizes AdamW~\cite{Kingma2014AdamAM} optimization with an initial learning rate of \(4 \times 10^{-4}\). We employ mixed-precision training with dynamic loss scaling, gradient clipping at $\|\nabla\|_2=0.1$, and weight decay regularization ($\lambda=5\times10^{-4}$). Distributed training executes on NVIDIA A100 clusters for 40K iterations - 32 GPUs for 500M/700M models (16 samples/GPU), 64 GPUs for the 1B parameter variant (8 samples/GPU). The total training times reach 78, 112, and 189 hours, respectively. During training, we randomly sample a source view image and four target view images from a video sequence.  

\begin{figure*}[tb]
    \centering
    \includegraphics[width=0.95\textwidth]{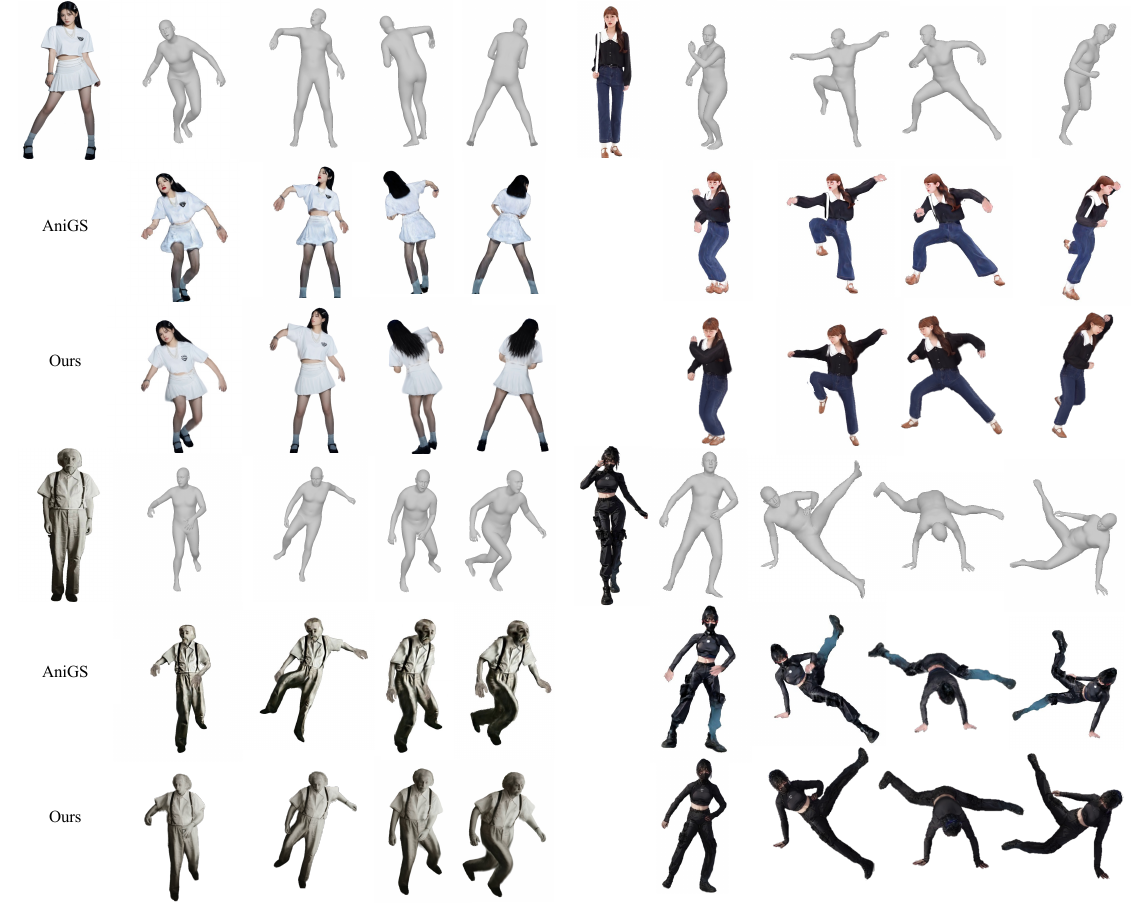}
    \\
    \vspace{-0.5em}
    \caption{
        Single-view animatable human reconstruction comparisons on in-the-wild sequences. 
        \MethodName produces more accurate and photorealistic animation results than the baseline methods.
        Note that the results of AniGS are not faithful to the input images.
    }
    \label{fig:qualitative_on_animate}
    \vspace{-1em}
\end{figure*}

\subsection{Comparison with Existing Methods}

\paragraph{Single-Image Human Reconstruction} We evaluate \MethodName against four baseline methods for single-view image human reconstruction. GTA~\cite{zhang2023globalcorrelated} and SIFU~\cite{zhang2024sifu} employ recursive optimization loops and pixel-aligned feature extraction, respectively, focusing on geometric refinement through successive approximation steps. PSHuman~\cite{li2024pshuman} employs multi-view diffusion in conjunction with local ID diffusion to boost the quality of facial features in multi-view human RGB and normal images, which is subsequently followed by multi-view mesh reconstruction.
 DreamGaussian~\cite{tang2023dreamgaussian} leverages score distillation sampling (SDS)~\cite{poole2022dreamfusion} from 2D diffusion models to distill 3D representations. While their progressive Gaussian densification strategy reduces convergence time to approximately 2 minutes per asset, this remains orders of magnitude slower than real-time requirements.

\Tref{tab:static_rec} compares the quantitative results on 200 synthetic datasets among baseline methods on four metrics: PSNR, SSIM, LPIPS, and Face Consistency~(FC) measured via L2 distance in the ArcFace~\cite{deng2019arcface} embedding space. Notably, for a fair comparison, we report metrics of the model trained on the same synthetic dataset as baseline methods.

With respect to qualitative results, as illustrated in \fref{fig:qualitative_on_static},  the visual comparisons highlight our method's ability to maintain input-aligned features while suppressing common artifacts such as over-smoothing.

\begin{table}[tb]\centering
    \caption{Evaluation of static 3D reconstruction on synthtic data. * indicates this model only trains on synthetic data.
    }
    \label{tab:static_rec}

    \resizebox{0.43\textwidth}{!}{
    \large
    \begin{tabular}{*{10}{c}}
        \toprule
       Methods & PSNR $\uparrow$ &  SSIM $\uparrow$ & LPIPS $\downarrow$ & FC $\downarrow$ \\
        \midrule
        GTA~\cite{zhang2023globalcorrelated} & 17.025 &  0.919 & 0.087 & 0.051 &  \\
        SIFu~\cite{zhang2024sifu} & 16.681 & 0.917 & 0.093 & 0.060 &  \\
        PSHuman~\cite{li2024pshuman} & 17.556 & 0.921 & 0.076 & 0.037 &  \\
        DreamGaussian~\cite{pan2024humansplat} & 18.544 & 0.917 & 0.075 & 0.056 &  \\
        \midrule
        % LHM-0.5B & \Scnd{25.144} &  \Scnd{0.942}  & 0.046 & 0.036 \\
        % LHM-0.7B & 25.112 &  0.940   & \Scnd{0.044} & \Scnd{0.035} \\
        % LHM-1B & \Frst{25.163} &  \Frst{0.944}  & \Frst{0.041} & \Frst{0.030} \\
        LHM-0.5B* & \textbf{25.183} & \textbf{0.951} & \textbf{0.029} & \textbf{0.035} \\ 
        \bottomrule
    \end{tabular}
    }

    \vspace{-1em}
\end{table}

\paragraph{Single-Image Animatable Human Reconstruction} We assess \MethodName against two baseline approaches for reconstructing animatable humans from a single-view image. The first baseline is En3D~\cite{men2024en3denhancedgenerativemodel}, which generates 3D human models in canonical space using physics-based 2D data alongside normal-constrained sculpting techniques. The second baseline, AniGS~\cite{qiu2024AniGS}, utilizes multi-view diffusion models to create canonical human images and employs 4D Gaussian splatting~(4DGS) optimization to address inconsistencies in different views.

For our evaluation, we utilize 200 in-the-wild video sequences drawn from the validation subset of our dataset. Specifically, we take the first front view image of each video as input and compare the synthesized animations against the corresponding ground-truth sequences using foreground segmentation masks. As shown in \Tref{tab:ani_rec}, our method surpasses the baseline approaches, demonstrating superior rendering quality in the animation sequences. In comparison to the best baseline method, AniGS, our approach achieves performance gains of 3.322, 0.059, 0.063, and 0.018 in PSNR, SSIM, LIPIS, and FC metrics, respectively. As illustrated in \fref{fig:qualitative_on_animate}, our method yields more accurate and photorealistic animation results compared to the baseline techniques. Additional results can be found in the supplementary material.

\begin{table}[tb]\centering
    \caption{Human animation results on in-the-wild video dataset. %\Frst{Red} text indicates the best and \Scnd{blue} text indicates the second best result, respectively. FC means face consistency.
    }
    \label{tab:ani_rec}
    
    \resizebox{0.48\textwidth}{!}{
    \large
    \begin{tabular}{*{10}{c}}
        \toprule
       Methods & PSNR $\uparrow$ &  SSIM $\uparrow$ & LPIPS $\downarrow$ & FC $\downarrow$ & Time$\downarrow$ & Memory $\downarrow$ \\
        \midrule
        En3D~\cite{men2024en3denhancedgenerativemodel} & 15.231 & 0.734  & 0.172 & 0.058 & 5 minutes & 32~GB \\
        AniGS~\cite{qiu2024AniGS} & 18.681 & 0.871 & 0.103 & 0.053 & 15 minutes & 24~GB \\
        \midrule
        LHM-0.5B & 21.648 &  0.924  & 0.044 & 0.042 & \Frst{2.01~seconds} & \Frst{18~GB} \\
        LHM-0.7B & \Scnd{21.879} & \Scnd{0.930}  & \Frst{0.039} & \Scnd{0.039} & \Scnd{4.13~seconds} & \Scnd{21~GB}\\
        LHM-1B & \Frst{22.003} & \Frst{0.930}  & \Scnd{0.040} & \Frst{0.035} & 6.57~seconds & 24~GB &\\
        \bottomrule
    \end{tabular}
    }

%\end{table}
%\begin{table}[tb]

    \centering
    \includegraphics[width=0.35\textwidth]{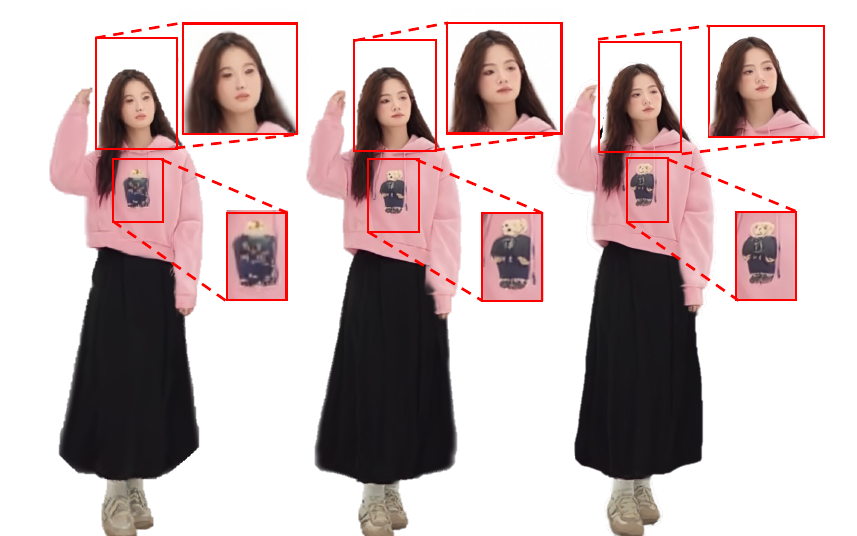}
    \\
    \makebox[0.12\textwidth]{\scriptsize (a) w/ MM-Transformer}
    \makebox[0.12\textwidth]{\scriptsize (b) LHM-0.5B}
    \makebox[0.12\textwidth]{\scriptsize (c) LHM-1B} 
    \captionof{figure}{Ablation study on model design and parameters.}
     %The \fcolorbox{red}{white}{red boxes} highlights the different areas.
    \label{fig:model_capacity}
    \vspace{-1em}
\end{table}

\subsection{Ablation Study}

\paragraph{Model Parameter Scalability}
To verify the scalability of our \MethodName, we train variant models with increasing parameter numbers by scaling the layer numbers.
\Tref{tab:ani_rec} compares performance across various model capacities. Our experiments indicate that increasing the number of model parameters correlates with improved performance. \Fref{fig:model_capacity} presents the comparison between LHM-0.5B and LHM-1B, where the larger model achieves more accurate reconstruction, especially in the face regions.

\paragraph{Dataset Scalability} To evaluate data scalability, we conduct controlled experiments using stratified random subsets (10K, 50K, 100K) from the original video training dataset of 300K. \Tref{tab:ablation_for_data} illustrates that using only the synthetic dataset results in poor model generalization. Incorporating an in-the-wild dataset significantly enhances the model's generality and performance on in-the-wild tests. Moreover, larger dataset sizes yield improved model results, although the rate of performance improvement diminishes as the dataset size increases. \Fref{fig:dataset_scalibity} showcases the ablation study on dataset scalability.

\begin{table}[t] \begin{center}
    \caption{Quantitative results on in-the-wild video dataset. %\Frst{Red} text indicates the best and \Scnd{blue} text indicates the second best result, respectively. FC means face consistency.
    }
    \label{tab:ablation_for_data}
    %\begin{table}[tb]\centering
%    \caption{Ablation for training with in-the-wild video datasets.}
%    \label{tab:ablation_for_data}
    \resizebox{0.45\textwidth}{!}{
    \large
    \begin{tabular}{*{10}{c}}
        \toprule
       Methods & PSNR $\uparrow$ &  SSIM $\uparrow$ & LPIPS $\downarrow$ & FC $\downarrow$  \\
        \midrule
        LHM-0.5B + Synthetic Data & 19.753 & 0.904 & 0.060 & 0.057 &  \\
        LHM-0.5B + 10K Videos  & 20.692 & 0.911 & 0.052 & 0.048 &  \\
        LHM-0.5B + 50K Videos  & 21.108 &  0.915  & 0.050 & 0.043 \\
        LHM-0.5B + 100K Videos  & 21.429 &  0.920  & 0.049 & 0.045 \\
        LHM-0.5B + All  &\textbf{21.648 }& \textbf{ 0.924 } & \textbf{0.044 }& \textbf{0.042} \\
        \bottomrule
    \end{tabular}
    }
    %\end{table}
    \end{center}
%\end{table}

%\begin{table}[tb]\centering
    \vspace{-1em}
    \caption{Ablation study for the transformer architecture.}
     %on in-the-wild video datasets
    \label{tab:ablation_for_model}
    %\begin{table}[tb]\centering
%    \caption{Ablation study for the architecture on in-the-wild video datasets.}
    %\label{tab:ablation_for_model}
    \resizebox{0.48\textwidth}{!}{
    \large
    \begin{tabular}{*{10}{c}}
        \toprule
       Methods & PSNR $\uparrow$ &  SSIM $\uparrow$ & LPIPS $\downarrow$ & FC $\downarrow$  \\
        \midrule
        LHM-0.5B w/ MM-Transformer & 20.072 & 0.907 & 0.100 & 0.053 &  \\
        LHM-0.5B w/o Shrinkage Regularization & 21.037 &  0.915 & 0.049 & \textbf{0.041} \\
        LHM-0.5B & \textbf{21.648} &  \textbf{0.924}  & \textbf{0.044} & 0.042 \\
        \bottomrule
    \end{tabular}
    }
%\end{table}
%\end{table}

%\begin{table}[tb]
\vspace{-0.0em}
    \centering
    \includegraphics[width=0.42\textwidth]{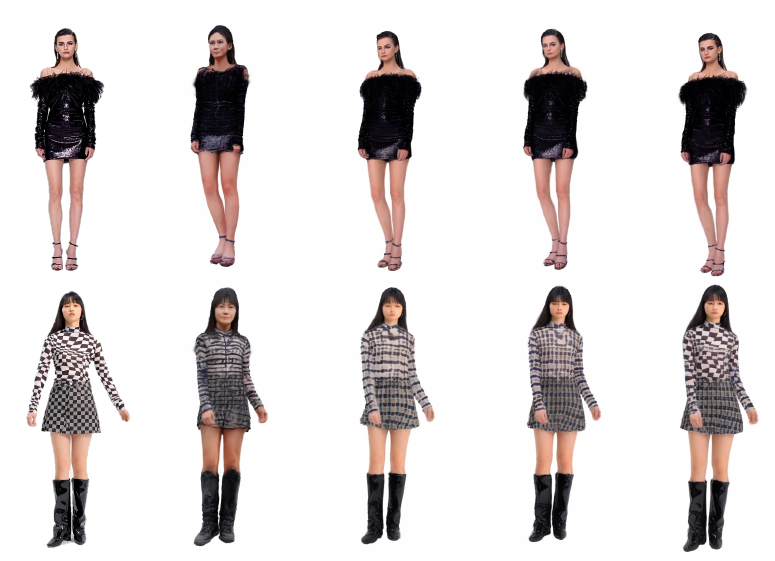}
    \\
    \vspace{-0.7em}
    \makebox[0.08\textwidth]{\scriptsize Input}
    \makebox[0.08\textwidth]{\scriptsize Synthetic Data}
    \makebox[0.08\textwidth]{\scriptsize 10K Videos} 
    \makebox[0.08\textwidth]{\scriptsize 100K Videos}
    \makebox[0.08\textwidth]{\scriptsize 300K Videos}
    \captionof{table}{Ablation study on dataset scalability.
    }
    \label{fig:dataset_scalibity}
    %\vspace{-2em}
\end{table}

\paragraph{Transformer Block Design} \Tref{tab:ablation_for_model} presents the quantitative results of our transformer block design. Compared to the vanilla MM-transformer block, our proposed transformer block demonstrates performance improvements of 1.576, 0.017, 0.056, and 0.011 in PSNR, SSIM, LIPIS, and FC metrics, respectively. Additionally, shrinkage regularization enhances the overall performance of our model, albeit with a slight reduction in face consistency. \Fref{fig:model_capacity} illustrates the qualitative results comparing the vanilla MM-transformer with our proposed BH-Transformer block.

\section{Conclusion}
\label{sec:Conclusion}

In this work, we introduce LHM, a feed-forward model for animatable 3D human reconstruction from a single image in seconds. 
Our approach leverages a multimodal transformer and a \headmodulefull scheme to effectively fuse 3D positional features and 2D image features via an attention mechanism, enabling joint reasoning across geometric and visual domains. Trained on a large-scale video dataset with an image reconstruction loss, our model exhibits strong generalization ability to diverse real-world scenarios. Extensive experiments on both synthetic and in-the-wild datasets demonstrate that LHM achieves state-of-the-art reconstruction accuracy, generalization, and animation consistency.

\paragraph{Limitations and Future Work} One limitation of our approach is that real-world video datasets often contain biased view distributions, with limited coverage of uncommon poses and extreme angles. This imbalance can affect the model’s ability to generalize to novel viewpoints. In future work, we aim to develop improved training strategies and curate a more diverse and comprehensive dataset to enhance robustness.

%In this work, we present a robust approach to generate ani- matable human avatars from a single image. We introduce a reference image-guided video generation model to pro- duce high-quality multi-view canonical human images and their corresponding normal maps. To handle view inconsis- tencies, we propose a 4D Gaussian Splatting (4DGS)-based method for reconstructing high-fidelity 3D avatars. Com- prehensive evaluation demonstrates that our method enables photorealistic, real-time animation of 3D human avatars from in-the-wild images.
%Limitations and Future Work While our method sup- ports real-time inference, it still requires several minutes to optimize an animatable avatar. In future work, we aim to explore feed-forward 3D reconstruction techniques that are robust to multi-view inconsistencies.

{\small
\bibliographystyle{ieeenat_fullname}
\bibliography{ref}

\begin{thebibliography}{70}
\providecommand{\natexlab}[1]{#1}
\providecommand{\url}[1]{\texttt{#1}}
\expandafter\ifx\csname urlstyle\endcsname\relax
  \providecommand{\doi}[1]{doi: #1}\else
  \providecommand{\doi}{doi: \begingroup \urlstyle{rm}\Url}\fi

\bibitem[Alldieck et~al.(2018{\natexlab{a}})Alldieck, Magnor, Xu, Theobalt, and Pons-Moll]{alldieck2018detailedhumanavatarsmonocular}
Thiemo Alldieck, Marcus Magnor, Weipeng Xu, Christian Theobalt, and Gerard Pons-Moll.
\newblock Detailed human avatars from monocular video, 2018{\natexlab{a}}.

\bibitem[Alldieck et~al.(2018{\natexlab{b}})Alldieck, Magnor, Xu, Theobalt, and Pons-Moll]{alldieck2018videobasedreconstruction3d}
Thiemo Alldieck, Marcus Magnor, Weipeng Xu, Christian Theobalt, and Gerard Pons-Moll.
\newblock Video based reconstruction of 3d people models, 2018{\natexlab{b}}.

\bibitem[Alldieck et~al.(2019)Alldieck, Pons-Moll, Theobalt, and Magnor]{alldieck2019tex2shapedetailedhumanbody}
Thiemo Alldieck, Gerard Pons-Moll, Christian Theobalt, and Marcus Magnor.
\newblock Tex2shape: Detailed full human body geometry from a single image, 2019.

\bibitem[Baradel* et~al.(2024)Baradel*, Armando, Galaaoui, Br{\'e}gier, Weinzaepfel, Rogez, and Lucas*]{multi-hmr2024}
Fabien Baradel*, Matthieu Armando, Salma Galaaoui, Romain Br{\'e}gier, Philippe Weinzaepfel, Gr{\'e}gory Rogez, and Thomas Lucas*.
\newblock Multi-hmr: Multi-person whole-body human mesh recovery in a single shot.
\newblock In \emph{ECCV}, 2024.

\bibitem[Cao et~al.(2022)Cao, Chen, Han, Yang, and Wong]{cao2022jiff}
Yukang Cao, Guanying Chen, Kai Han, Wenqi Yang, and Kwan-Yee~K Wong.
\newblock Jiff: Jointly-aligned implicit face function for high quality single view clothed human reconstruction.
\newblock In \emph{Proceedings of the IEEE/CVF Conference on Computer Vision and Pattern Recognition}, pages 2729--2739, 2022.

\bibitem[Cao et~al.(2024)Cao, Cao, Han, Shan, and Wong]{cao2024dreamavatar}
Yukang Cao, Yan-Pei Cao, Kai Han, Ying Shan, and Kwan-Yee~K Wong.
\newblock Dreamavatar: Text-and-shape guided 3d human avatar generation via diffusion models.
\newblock In \emph{Proceedings of the IEEE/CVF Conference on Computer Vision and Pattern Recognition}, pages 958--968, 2024.

\bibitem[Chen et~al.(2024{\natexlab{a}})Chen, Chen, Ye, ang Gao, Chen, Fan, and Zhao]{chen2024ultramansingleimage3d}
Mingjin Chen, Junhao Chen, Xiaojun Ye, Huan ang Gao, Xiaoxue Chen, Zhaoxin Fan, and Hao Zhao.
\newblock Ultraman: Single image 3d human reconstruction with ultra speed and detail, 2024{\natexlab{a}}.

\bibitem[Chen et~al.(2024{\natexlab{b}})Chen, Zheng, Li, Xu, and Liu]{chen2024meshavatar}
Yushuo Chen, Zerong Zheng, Zhe Li, Chao Xu, and Yebin Liu.
\newblock Meshavatar: Learning high-quality triangular human avatars from multi-view videos, 2024{\natexlab{b}}.

\bibitem[Choutas et~al.(2022)Choutas, Muller, Huang, Tang, Tzionas, and Black]{choutas2022accurate3dbodyshape}
Vasileios Choutas, Lea Muller, Chun-Hao~P. Huang, Siyu Tang, Dimitrios Tzionas, and Michael~J. Black.
\newblock Accurate 3d body shape regression using metric and semantic attributes, 2022.

\bibitem[Deng et~al.(2019)Deng, Guo, Xue, and Zafeiriou]{deng2019arcface}
Jiankang Deng, Jia Guo, Niannan Xue, and Stefanos Zafeiriou.
\newblock Arcface: Additive angular margin loss for deep face recognition.
\newblock In \emph{CVPR}, pages 4690--4699, 2019.

\bibitem[Esser et~al.(2024)Esser, Kulal, Blattmann, Entezari, M{\"u}ller, Saini, Levi, Lorenz, Sauer, Boesel, et~al.]{esser2024scaling}
Patrick Esser, Sumith Kulal, Andreas Blattmann, Rahim Entezari, Jonas M{\"u}ller, Harry Saini, Yam Levi, Dominik Lorenz, Axel Sauer, Frederic Boesel, et~al.
\newblock Scaling rectified flow transformers for high-resolution image synthesis.
\newblock In \emph{ICML}, 2024.

\bibitem[Han et~al.(2023)Han, Park, Yoon, Kang, Park, and Jeon]{han2023highfidelity3dhumandigitization}
Sang-Hun Han, Min-Gyu Park, Ju~Hong Yoon, Ju-Mi Kang, Young-Jae Park, and Hae-Gon Jeon.
\newblock High-fidelity 3d human digitization from single 2k resolution images.
\newblock In \emph{CVPR}, 2023.

\bibitem[He et~al.(2022{\natexlab{a}})He, Chen, Xie, Li, Doll{\'a}r, and Girshick]{MaskedAutoencoders2022}
Kaiming He, Xinlei Chen, Saining Xie, Yanghao Li, Piotr Doll{\'a}r, and Ross Girshick.
\newblock Masked autoencoders are scalable vision learners.
\newblock In \emph{CVPR}, 2022{\natexlab{a}}.

\bibitem[He et~al.(2022{\natexlab{b}})He, Xu, Saito, Soatto, and Tung]{he2022archanimationreadyclothedhuman}
Tong He, Yuanlu Xu, Shunsuke Saito, Stefano Soatto, and Tony Tung.
\newblock Arch++: Animation-ready clothed human reconstruction revisited, 2022{\natexlab{b}}.

\bibitem[He et~al.(2024)He, Li, Kang, Ye, Zhang, Chen, Gao, Zhang, Wu, and Zhuang]{he2024magicmangenerativenovelview}
Xu He, Xiaoyu Li, Di Kang, Jiangnan Ye, Chaopeng Zhang, Liyang Chen, Xiangjun Gao, Han Zhang, Zhiyong Wu, and Haolin Zhuang.
\newblock Magicman: Generative novel view synthesis of humans with 3d-aware diffusion and iterative refinement, 2024.

\bibitem[Hong et~al.(2023)Hong, Zhang, Gu, Bi, Zhou, Liu, Liu, Sunkavalli, Bui, and Tan]{hong2023lrm}
Yicong Hong, Kai Zhang, Jiuxiang Gu, Sai Bi, Yang Zhou, Difan Liu, Feng Liu, Kalyan Sunkavalli, Trung Bui, and Hao Tan.
\newblock Lrm: Large reconstruction model for single image to 3d.
\newblock \emph{arXiv preprint arXiv:2311.04400}, 2023.

\bibitem[Hu(2024)]{hu2024animate}
Li Hu.
\newblock Animate anyone: Consistent and controllable image-to-video synthesis for character animation.
\newblock In \emph{Proceedings of the IEEE/CVF Conference on Computer Vision and Pattern Recognition}, pages 8153--8163, 2024.

\bibitem[Hu and Liu(2024)]{hu2023gauhuman}
Shoukang Hu and Ziwei Liu.
\newblock Gauhuman: Articulated gaussian splatting from monocular human videos.
\newblock In \emph{CVPR}, 2024.

\bibitem[Huang et~al.(2023)Huang, Shao, Zhang, Zhang, Feng, Liu, and Wang]{huang2023humannorm}
Xin Huang, Ruizhi Shao, Qi Zhang, Hongwen Zhang, Ying Feng, Yebin Liu, and Qing Wang.
\newblock Humannorm: Learning normal diffusion model for high-quality and realistic 3d human generation.
\newblock \emph{arXiv preprint arXiv:2310.01406}, 2023.

\bibitem[Huang et~al.(2024{\natexlab{a}})Huang, Wang, Zeng, Cao, Qi, Shi, Zha, and Zhang]{huang2024dreamwaltz}
Yukun Huang, Jianan Wang, Ailing Zeng, He Cao, Xianbiao Qi, Yukai Shi, Zheng-Jun Zha, and Lei Zhang.
\newblock Dreamwaltz: Make a scene with complex 3d animatable avatars.
\newblock \emph{Advances in Neural Information Processing Systems}, 36, 2024{\natexlab{a}}.

\bibitem[Huang et~al.(2024{\natexlab{b}})Huang, Yi, Xiu, Liao, Tang, Cai, and Thies]{huang2024tech}
Yangyi Huang, Hongwei Yi, Yuliang Xiu, Tingting Liao, Jiaxiang Tang, Deng Cai, and Justus Thies.
\newblock {TeCH: Text-guided Reconstruction of Lifelike Clothed Humans}.
\newblock In \emph{International Conference on 3D Vision (3DV)}, 2024{\natexlab{b}}.

\bibitem[Huang et~al.(2020)Huang, Xu, Lassner, Li, and Tung]{huang2020archanimatablereconstructionclothed}
Zeng Huang, Yuanlu Xu, Christoph Lassner, Hao Li, and Tony Tung.
\newblock Arch: Animatable reconstruction of clothed humans, 2020.

\bibitem[Jiang et~al.(2022)Jiang, Hong, Bao, and Zhang]{jiang2022selfrecon}
Boyi Jiang, Yang Hong, Hujun Bao, and Juyong Zhang.
\newblock Selfrecon: Self reconstruction your digital avatar from monocular video.
\newblock In \emph{CVPR}, 2022.

\bibitem[Kanazawa et~al.(2018)Kanazawa, Black, Jacobs, and Malik]{kanazawa2018endtoendrecoveryhumanshape}
Angjoo Kanazawa, Michael~J. Black, David~W. Jacobs, and Jitendra Malik.
\newblock End-to-end recovery of human shape and pose, 2018.

\bibitem[Kerbl et~al.(2023)Kerbl, Kopanas, Leimk{\"u}hler, and Drettakis]{kerbl3Dgaussians}
Bernhard Kerbl, Georgios Kopanas, Thomas Leimk{\"u}hler, and George Drettakis.
\newblock 3d gaussian splatting for real-time radiance field rendering.
\newblock \emph{TOG}, 2023.

\bibitem[Khirodkar et~al.(2024)Khirodkar, Bagautdinov, Martinez, Zhaoen, James, Selednik, Anderson, and Saito]{khirodkar2024sapiens}
Rawal Khirodkar, Timur Bagautdinov, Julieta Martinez, Su Zhaoen, Austin James, Peter Selednik, Stuart Anderson, and Shunsuke Saito.
\newblock Sapiens: Foundation for human vision models.
\newblock In \emph{European Conference on Computer Vision}, pages 206--228. Springer, 2024.

\bibitem[Kingma and Ba(2014)]{Kingma2014AdamAM}
Diederik~P. Kingma and Jimmy Ba.
\newblock Adam: A method for stochastic optimization.
\newblock \emph{CoRR}, abs/1412.6980, 2014.

\bibitem[Kolotouros et~al.(2024)Kolotouros, Alldieck, Zanfir, Bazavan, Fieraru, and Sminchisescu]{kolotouros2024dreamhuman}
Nikos Kolotouros, Thiemo Alldieck, Andrei Zanfir, Eduard Bazavan, Mihai Fieraru, and Cristian Sminchisescu.
\newblock Dreamhuman: Animatable 3d avatars from text.
\newblock \emph{Advances in Neural Information Processing Systems}, 36, 2024.

\bibitem[Li et~al.(2024{\natexlab{a}})Li, Zheng, Liu, Yu, Li, Qi, Li, Chi, Xia, Xue, et~al.]{li2024pshuman}
Peng Li, Wangguandong Zheng, Yuan Liu, Tao Yu, Yangguang Li, Xingqun Qi, Mengfei Li, Xiaowei Chi, Siyu Xia, Wei Xue, et~al.
\newblock Pshuman: Photorealistic single-view human reconstruction using cross-scale diffusion.
\newblock \emph{arXiv preprint arXiv:2409.10141}, 2024{\natexlab{a}}.

\bibitem[Li et~al.(2024{\natexlab{b}})Li, Zheng, Wang, and Liu]{li2024animatable}
Zhe Li, Zerong Zheng, Lizhen Wang, and Yebin Liu.
\newblock Animatable gaussians: Learning pose-dependent gaussian maps for high-fidelity human avatar modeling.
\newblock In \emph{Proceedings of the IEEE/CVF Conference on Computer Vision and Pattern Recognition}, pages 19711--19722, 2024{\natexlab{b}}.

\bibitem[Li et~al.(2024{\natexlab{c}})Li, Zheng, Wang, and Liu]{li2024animatablegaussians}
Zhe Li, Zerong Zheng, Lizhen Wang, and Yebin Liu.
\newblock Animatable gaussians: Learning pose-dependent gaussian maps for high-fidelity human avatar modeling.
\newblock In \emph{CVPR}, 2024{\natexlab{c}}.

\bibitem[Lin et~al.(2025)Lin, Jiang, Yang, Zheng, and Liang]{lin2025omnihuman}
Gaojie Lin, Jianwen Jiang, Jiaqi Yang, Zerong Zheng, and Chao Liang.
\newblock Omnihuman-1: Rethinking the scaling-up of one-stage conditioned human animation models.
\newblock \emph{arXiv preprint arXiv:2502.01061}, 2025.

\bibitem[Liu et~al.(2016)Liu, Luo, Qiu, Wang, and Tang]{liuLQWTcvpr16DeepFashion}
Ziwei Liu, Ping Luo, Shi Qiu, Xiaogang Wang, and Xiaoou Tang.
\newblock Deepfashion: Powering robust clothes recognition and retrieval with rich annotations.
\newblock In \emph{CVPR}, 2016.

\bibitem[Loper et~al.(2015)Loper, Mahmood, Romero, Pons-Moll, and Black]{loper2015smpl}
Matthew Loper, Naureen Mahmood, Javier Romero, Gerard Pons-Moll, and Michael~J Black.
\newblock Smpl: a skinned multi-person linear model.
\newblock \emph{TOG}, 34\penalty0 (6):\penalty0 1--16, 2015.

\bibitem[Lu et~al.(2025)Lu, Dong, Kwon, Zhao, Dai, and De~la Torre]{lu2025gas}
Yixing Lu, Junting Dong, Youngjoong Kwon, Qin Zhao, Bo Dai, and Fernando De~la Torre.
\newblock Gas: Generative avatar synthesis from a single image.
\newblock \emph{arXiv preprint arXiv:2502.06957}, 2025.

\bibitem[Men et~al.(2024{\natexlab{a}})Men, Lei, Yao, Cui, Lian, and Xie]{men2024en3denhancedgenerativemodel}
Yifang Men, Biwen Lei, Yuan Yao, Miaomiao Cui, Zhouhui Lian, and Xuansong Xie.
\newblock En3d: An enhanced generative model for sculpting 3d humans from 2d synthetic data, 2024{\natexlab{a}}.

\bibitem[Men et~al.(2024{\natexlab{b}})Men, Yao, Cui, and Bo]{men2024mimo}
Yifang Men, Yuan Yao, Miaomiao Cui, and Liefeng Bo.
\newblock Mimo: Controllable character video synthesis with spatial decomposed modeling.
\newblock \emph{arXiv preprint arXiv:2409.16160}, 2024{\natexlab{b}}.

\bibitem[Mildenhall et~al.(2020)Mildenhall, Srinivasan, Tancik, Barron, Ramamoorthi, and Ng]{mildenhall2020nerf}
Ben Mildenhall, Pratul~P Srinivasan, Matthew Tancik, Jonathan~T Barron, Ravi Ramamoorthi, and Ren Ng.
\newblock {NeRF}: Representing scenes as neural radiance fields for view synthesis.
\newblock In \emph{ECCV}, pages 405--421, 2020.

\bibitem[Moon et~al.(2024)Moon, Shiratori, and Saito]{moon2024exavatar}
Gyeongsik Moon, Takaaki Shiratori, and Shunsuke Saito.
\newblock Expressive whole-body 3d gaussian avatar.
\newblock In \emph{ECCV}, 2024.

\bibitem[Oquab et~al.(2023)Oquab, Darcet, Moutakanni, Vo, Szafraniec, Khalidov, Fernandez, Haziza, Massa, El-Nouby, Howes, Huang, Xu, Sharma, Li, Galuba, Rabbat, Assran, Ballas, Synnaeve, Misra, Jegou, Mairal, Labatut, Joulin, and Bojanowski]{oquab2023dinov2}
Maxime Oquab, Timothée Darcet, Theo Moutakanni, Huy~V. Vo, Marc Szafraniec, Vasil Khalidov, Pierre Fernandez, Daniel Haziza, Francisco Massa, Alaaeldin El-Nouby, Russell Howes, Po-Yao Huang, Hu Xu, Vasu Sharma, Shang-Wen Li, Wojciech Galuba, Mike Rabbat, Mido Assran, Nicolas Ballas, Gabriel Synnaeve, Ishan Misra, Herve Jegou, Julien Mairal, Patrick Labatut, Armand Joulin, and Piotr Bojanowski.
\newblock Dinov2: Learning robust visual features without supervision, 2023.

\bibitem[Pan et~al.(2024)Pan, Su, Lin, Fan, Zhang, Li, Shen, Mu, and Liu]{pan2024humansplat}
Panwang Pan, Zhuo Su, Chenguo Lin, Zhen Fan, Yongjie Zhang, Zeming Li, Tingting Shen, Yadong Mu, and Yebin Liu.
\newblock Humansplat: Generalizable single-image human gaussian splatting with structure priors.
\newblock \emph{arXiv preprint arXiv:2406.12459}, 2024.

\bibitem[Pan et~al.(2025)Pan, Su, Lin, Fan, Zhang, Li, Shen, Mu, and Liu]{pan2025humansplat}
Panwang Pan, Zhuo Su, Chenguo Lin, Zhen Fan, Yongjie Zhang, Zeming Li, Tingting Shen, Yadong Mu, and Yebin Liu.
\newblock Humansplat: Generalizable single-image human gaussian splatting with structure priors.
\newblock \emph{Advances in Neural Information Processing Systems}, 37:\penalty0 74383--74410, 2025.

\bibitem[Pang et~al.(2024)Pang, Liu, Cai, Yang, Zhang, and Liu]{Pang2024Disco4DD4}
Hui~En Pang, Shuai Liu, Zhongang Cai, Lei Yang, Tianwei Zhang, and Ziwei Liu.
\newblock Disco4d: Disentangled 4d human generation and animation from a single image.
\newblock \emph{ArXiv}, abs/2409.17280, 2024.

\bibitem[Pavlakos et~al.(2019)Pavlakos, Choutas, Ghorbani, Bolkart, Osman, Tzionas, and Black]{smplx:2019}
Georgios Pavlakos, Vasileios Choutas, Nima Ghorbani, Timo Bolkart, Ahmed~AA Osman, Dimitrios Tzionas, and Michael~J Black.
\newblock Expressive body capture: 3d hands, face, and body from a single image.
\newblock In \emph{CVPR}, 2019.

\bibitem[Peng et~al.(2024)Peng, Zhang, Guo, Cao, and Hu]{peng2024charactergen}
Hao-Yang Peng, Jia-Peng Zhang, Meng-Hao Guo, Yan-Pei Cao, and Shi-Min Hu.
\newblock Charactergen: Efficient 3d character generation from single images with multi-view pose canonicalization.
\newblock \emph{ACM Transactions on Graphics (TOG)}, 43\penalty0 (4):\penalty0 1--13, 2024.

\bibitem[Peng et~al.(2021)Peng, Dong, Wang, Zhang, Shuai, Zhou, and Bao]{peng2021animatable}
Sida Peng, Junting Dong, Qianqian Wang, Shangzhan Zhang, Qing Shuai, Xiaowei Zhou, and Hujun Bao.
\newblock Animatable neural radiance fields for modeling dynamic human bodies.
\newblock In \emph{Proceedings of the IEEE/CVF International Conference on Computer Vision}, pages 14314--14323, 2021.

\bibitem[Poole et~al.(2023)Poole, Jain, Barron, and Mildenhall]{poole2022dreamfusion}
Ben Poole, Ajay Jain, Jonathan~T Barron, and Ben Mildenhall.
\newblock Dreamfusion: Text-to-3d using 2d diffusion.
\newblock In \emph{ICLR}, 2023.

\bibitem[Qiu and Chen(2023)]{qiu2023recmv}
Lingteng Qiu and Guanying Chen.
\newblock Rec-mv: Reconstructing 3d dynamic cloth from monocular videos.
\newblock In \emph{CVPR}, 2023.

\bibitem[Qiu et~al.(2025)Qiu, Zhu, Zuo, Gu, Dong, Zhang, Xu, Li, Yuan, Bo, et~al.]{qiu2024AniGS}
Lingteng Qiu, Shenhao Zhu, Qi Zuo, Xiaodong Gu, Yuan Dong, Junfei Zhang, Chao Xu, Zhe Li, Weihao Yuan, Liefeng Bo, et~al.
\newblock Anigs: Animatable gaussian avatar from a single image with inconsistent gaussian reconstruction.
\newblock In \emph{CVPR}, 2025.

\bibitem[Saito et~al.(2019)Saito, Huang, Natsume, Morishima, Kanazawa, and Li]{saito2019pifu}
Shunsuke Saito, Zeng Huang, Ryota Natsume, Shigeo Morishima, Angjoo Kanazawa, and Hao Li.
\newblock Pifu: Pixel-aligned implicit function for high-resolution clothed human digitization.
\newblock In \emph{Proceedings of the IEEE/CVF international conference on computer vision}, pages 2304--2314, 2019.

\bibitem[Saito et~al.(2020)Saito, Simon, Saragih, and Joo]{saito2020pifuhd}
Shunsuke Saito, Tomas Simon, Jason Saragih, and Hanbyul Joo.
\newblock Pifuhd: Multi-level pixel-aligned implicit function for high-resolution 3d human digitization.
\newblock In \emph{Proceedings of the IEEE/CVF conference on computer vision and pattern recognition}, pages 84--93, 2020.

\bibitem[Shao et~al.(2024)Shao, Pang, Zheng, Sun, and Liu]{shao2024human4dit}
Ruizhi Shao, Youxin Pang, Zerong Zheng, Jingxiang Sun, and Yebin Liu.
\newblock Human4dit: 360-degree human video generation with 4d diffusion transformer.
\newblock \emph{TOG}, 43\penalty0 (6), 2024.

\bibitem[Tan et~al.(2025)Tan, Xiang, Tulsiani, Ramanan, and Yang]{tan2025dressrecon}
Jeff Tan, Donglai Xiang, Shubham Tulsiani, Deva Ramanan, and Gengshan Yang.
\newblock Dressrecon: Freeform 4d human reconstruction from monocular video.
\newblock In \emph{3DV}, 2025.

\bibitem[Tang et~al.(2024)Tang, Ren, Zhou, Liu, and Zeng]{tang2023dreamgaussian}
Jiaxiang Tang, Jiawei Ren, Hang Zhou, Ziwei Liu, and Gang Zeng.
\newblock Dreamgaussian: Generative gaussian splatting for efficient 3d content creation.
\newblock In \emph{ICLR}, 2024.

\bibitem[Tang et~al.(2025)Tang, Chen, Chen, Wang, Zeng, and Liu]{tang2025lgm}
Jiaxiang Tang, Zhaoxi Chen, Xiaokang Chen, Tengfei Wang, Gang Zeng, and Ziwei Liu.
\newblock Lgm: Large multi-view gaussian model for high-resolution 3d content creation.
\newblock In \emph{European Conference on Computer Vision}, pages 1--18. Springer, 2025.

\bibitem[Weng et~al.(2022)Weng, Curless, Srinivasan, Barron, and Kemelmacher-Shlizerman]{weng2022humannerf}
Chung-Yi Weng, Brian Curless, Pratul~P Srinivasan, Jonathan~T Barron, and Ira Kemelmacher-Shlizerman.
\newblock Humannerf: Free-viewpoint rendering of moving people from monocular video.
\newblock In \emph{CVPR}, 2022.

\bibitem[Weng et~al.(2024)Weng, Liu, Tan, Xu, Zhou, Yeung-Levy, and Yang]{weng2024template}
Zhenzhen Weng, Jingyuan Liu, Hao Tan, Zhan Xu, Yang Zhou, Serena Yeung-Levy, and Jimei Yang.
\newblock Template-free single-view 3d human digitalization with diffusion-guided lrm.
\newblock \emph{arXiv preprint arXiv:2401.12175}, 2024.

\bibitem[Xiong et~al.(2024)Xiong, Li, Liu, Liao, Hu, Zhu, Ning, Qiu, Wang, Wang, et~al.]{xiong2024mvhumannet}
Zhangyang Xiong, Chenghong Li, Kenkun Liu, Hongjie Liao, Jianqiao Hu, Junyi Zhu, Shuliang Ning, Lingteng Qiu, Chongjie Wang, Shijie Wang, et~al.
\newblock Mvhumannet: A large-scale dataset of multi-view daily dressing human captures.
\newblock In \emph{CVPR}, 2024.

\bibitem[Xiu et~al.(2023)Xiu, Yang, Cao, Tzionas, and Black]{xiu2023econexplicitclothedhumans}
Yuliang Xiu, Jinlong Yang, Xu Cao, Dimitrios Tzionas, and Michael~J. Black.
\newblock Econ: Explicit clothed humans optimized via normal integration, 2023.

\bibitem[Xiu et~al.(2024)Xiu, Ye, Liu, Tzionas, and Black]{xiu2024puzzleavatar}
Yuliang Xiu, Yufei Ye, Zhen Liu, Dimitrios Tzionas, and Michael~J Black.
\newblock Puzzleavatar: Assembling 3d avatars from personal albums.
\newblock \emph{TOG}, 2024.

\bibitem[Xu et~al.(2023)Xu, Zhang, Liew, Feng, and Shou]{XAGen2023}
Zhongcong Xu, Jianfeng Zhang, Junhao Liew, Jiashi Feng, and Mike~Zheng Shou.
\newblock Xagen: 3d expressive human avatars generation.
\newblock In \emph{NeurIPS}, 2023.

\bibitem[Yang et~al.(2024{\natexlab{a}})Yang, Huang, Chai, Jiang, and Hwang]{yang2024samurai}
Cheng-Yen Yang, Hsiang-Wei Huang, Wenhao Chai, Zhongyu Jiang, and Jenq-Neng Hwang.
\newblock Samurai: Adapting segment anything model for zero-shot visual tracking with motion-aware memory, 2024{\natexlab{a}}.

\bibitem[Yang et~al.(2024{\natexlab{b}})Yang, Chen, Gao, Wang, Han, and Wang]{bib:havefun}
Xihe Yang, Xingyu Chen, Daiheng Gao, Shaohui Wang, Xiaoguang Han, and Baoyuan Wang.
\newblock Have-fun: Human avatar reconstruction from few-shot unconstrained images.
\newblock In \emph{CVPR}, 2024{\natexlab{b}}.

\bibitem[Yu et~al.(2023)Yu, Cheng, Liu, Wu, and Lin]{yu2023monohuman}
Zhengming Yu, Wei Cheng, Xian Liu, Wayne Wu, and Kwan-Yee Lin.
\newblock Monohuman: Animatable human neural field from monocular video.
\newblock In \emph{CVPR}, 2023.

\bibitem[Zhang et~al.(2023)Zhang, Sun, Yang, Chen, and Yang]{zhang2023globalcorrelated}
Zechuan Zhang, Li Sun, Zongxin Yang, Ling Chen, and Yi Yang.
\newblock Global-correlated 3d-decoupling transformer for clothed avatar reconstruction.
\newblock In \emph{NeurIPS}, 2023.

\bibitem[Zhang et~al.(2024{\natexlab{a}})Zhang, Yang, and Yang]{zhang2024sifu}
Zechuan Zhang, Zongxin Yang, and Yi Yang.
\newblock Sifu: Side-view conditioned implicit function for real-world usable clothed human reconstruction.
\newblock In \emph{Proceedings of the IEEE/CVF Conference on Computer Vision and Pattern Recognition}, pages 9936--9947, 2024{\natexlab{a}}.

\bibitem[Zhang et~al.(2024{\natexlab{b}})Zhang, Yang, and Yang]{zhang2024sifusideviewconditionedimplicit}
Zechuan Zhang, Zongxin Yang, and Yi Yang.
\newblock Sifu: Side-view conditioned implicit function for real-world usable clothed human reconstruction, 2024{\natexlab{b}}.

\bibitem[Zheng et~al.(2020)Zheng, Yu, Liu, and Dai]{zheng2020pamirparametricmodelconditionedimplicit}
Zerong Zheng, Tao Yu, Yebin Liu, and Qionghai Dai.
\newblock Pamir: Parametric model-conditioned implicit representation for image-based human reconstruction, 2020.

\bibitem[Zhu et~al.(2024)Zhu, Chen, Dai, Xu, Cao, Yao, Zhu, and Zhu]{zhu2024champ}
Shenhao Zhu, Junming~Leo Chen, Zuozhuo Dai, Yinghui Xu, Xun Cao, Yao Yao, Hao Zhu, and Siyu Zhu.
\newblock Champ: Controllable and consistent human image animation with 3d parametric guidance.
\newblock \emph{arXiv preprint arXiv:2403.14781}, 2024.

\bibitem[Zhuang et~al.(2024)Zhuang, Lv, Wen, Shuai, Zeng, Zhu, Chen, Yang, Cao, and Liu]{zhuang2024idol}
Yiyu Zhuang, Jiaxi Lv, Hao Wen, Qing Shuai, Ailing Zeng, Hao Zhu, Shifeng Chen, Yujiu Yang, Xun Cao, and Wei Liu.
\newblock Idol: Instant photorealistic 3d human creation from a single image.
\newblock \emph{arXiv preprint arXiv:2412.14963}, 2024.

\end{thebibliography}
}
\appendix

\clearpage
\section{Demo Video}
Please kindly check the \href{https://www.youtube.com/watch?v=tivEpz_yiEo}{Demo Video} for animation results of the reconstructed 3D avatar.

\section{Details of the Multimodal Transformer }

Our Multimodal Body-Head Transformer (MBHT) is built on top of the recent Multimodal Transformers (MM-Transformer)~\cite{esser2024scaling}. 

\begin{figure}[htb]
    \centering
    \includegraphics[width=1\linewidth]{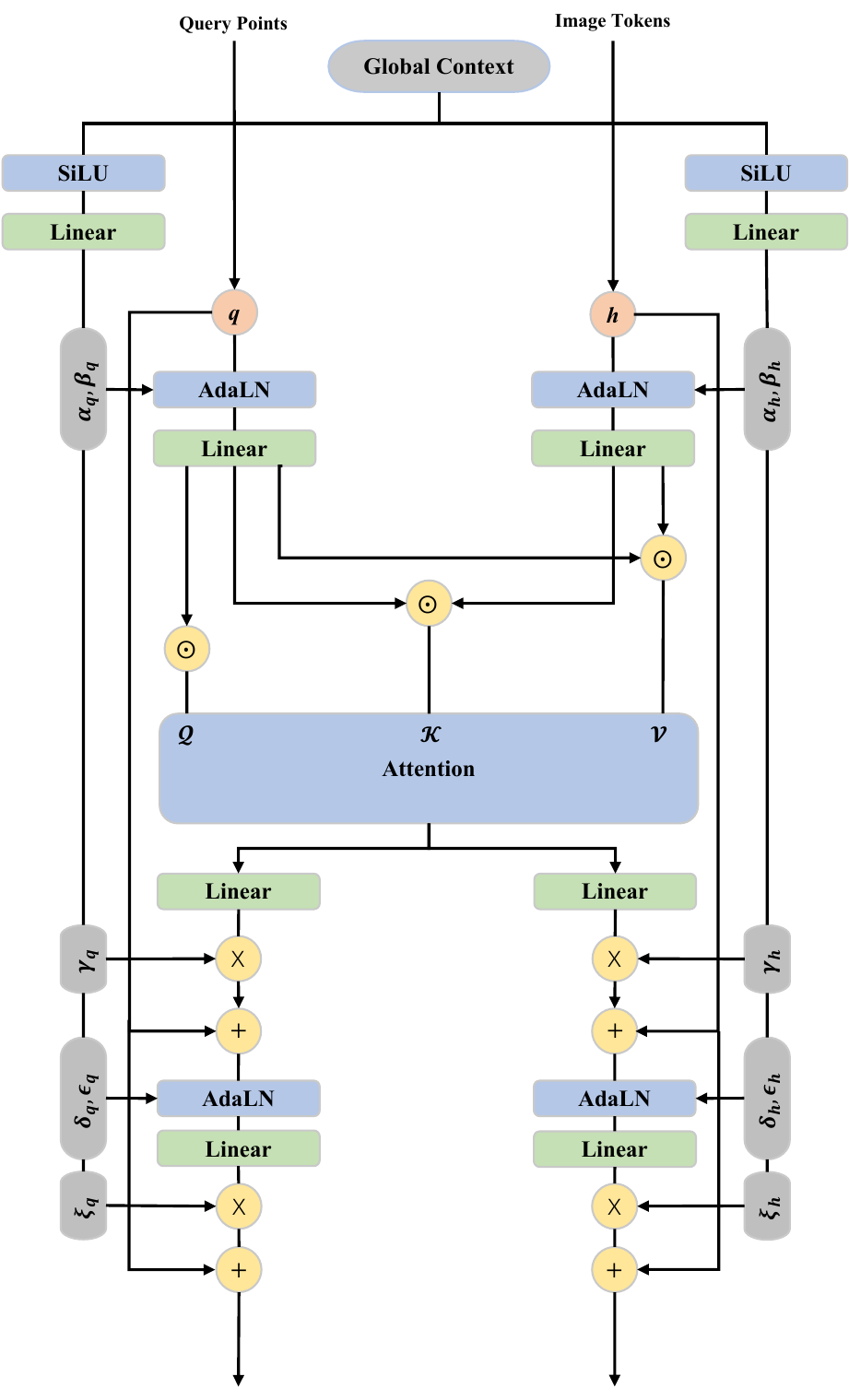}
    \caption{Detailed architecture of Multi-Modal Transformer~\cite{esser2024scaling}.}
    \label{fig:mm_transformer}
\end{figure}

The detailed architecture of MM-Transformer is summarized in \fref{fig:mm_transformer}. 
The 3D geometric body and head query tokens are fed as $q$ and semantic image feature tokens are fed as $h$. 
MM-Transformer aggregates both features by attention mechanism with Adaptive Layer Normalization modulation guided by the extracted global context features.

\section{Details of Head Feature Pyramid Encoding}
\begin{figure}[tb]
    \centering
    \includegraphics[width=0.5\textwidth]{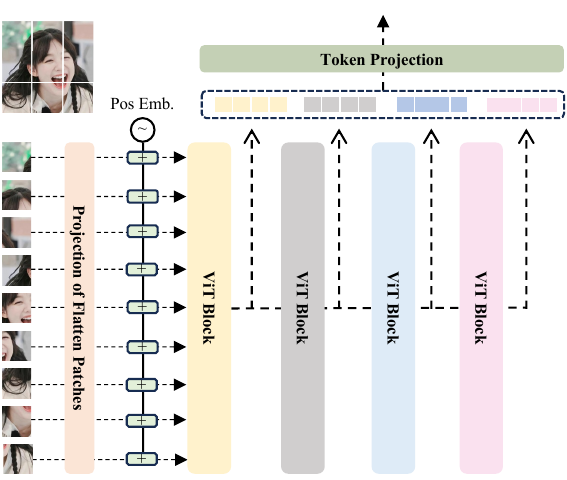}
    \caption{Architecture of our \HeadModule for multi-scale facial feature extraction}
    \label{fig:pyramid_head_encoder}
\end{figure}

Given that the human head occupies a relatively small area within the input image and is subject to spatial downsampling during the encoding process, essential facial details are frequently lost. To address this challenge, we introduce a \headmodulefull (\HeadModule) designed to aggregate multi-scale features of DINOv2~\cite{oquab2023dinov2}.
\Fref{fig:pyramid_head_encoder} illustrates the architecture of \HeadModule.

\section{Details of the Synthetic Training Dataset}
To address viewpoint bias in natural videos, we supplement training with synthetic human scans from three sources: (1)~2K2K dataset~\cite{han2023highfidelity3dhumandigitization} sampling 1,000 textured models, (2) Human4DiT~\cite{shao2024human4dit} sampling 4,324 textured characters, and (3) 400 commercial assets from RenderPeople, culminating in 5,724 high-fidelity 3D human scans. Following AniGS~\cite{qiu2024AniGS}'s multi-view rendering protocol, we generate 30 azimuthal views per model with uniform angular spacing (12\textdegree intervals) under HDRI lighting conditions.

\section{Effects of Canonical Space Regularization} 
\begin{figure}[tb]
    \centering
    \includegraphics[width=0.48\textwidth]{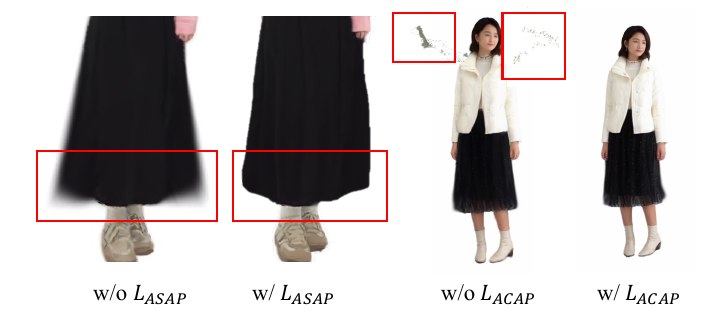}
    \caption{Ablation for canonical space shape regularization.}
    \label{fig:shape_reg}
\end{figure}

We conduct an ablation study to assess the impact of the canonical space regularization design.
\Fref{fig:shape_reg} shows that the \emph{as spherical as possible} loss $\mathcal{L}_{ASAP}$ is effective in reducing semi-transparent boundary artifacts caused by Gaussians with distorted shapes.

Without the \emph{as close as possible} loss $\mathcal{L}_{ACAP}$, the reconstruction results exhibit noticeable floating points around the human.
These results clearly demonstrate the effectiveness of the proposed canonical space regularization losses.

\section{More Results}

\begin{figure*}[tb] \centering
    \includegraphics[width=\textwidth]{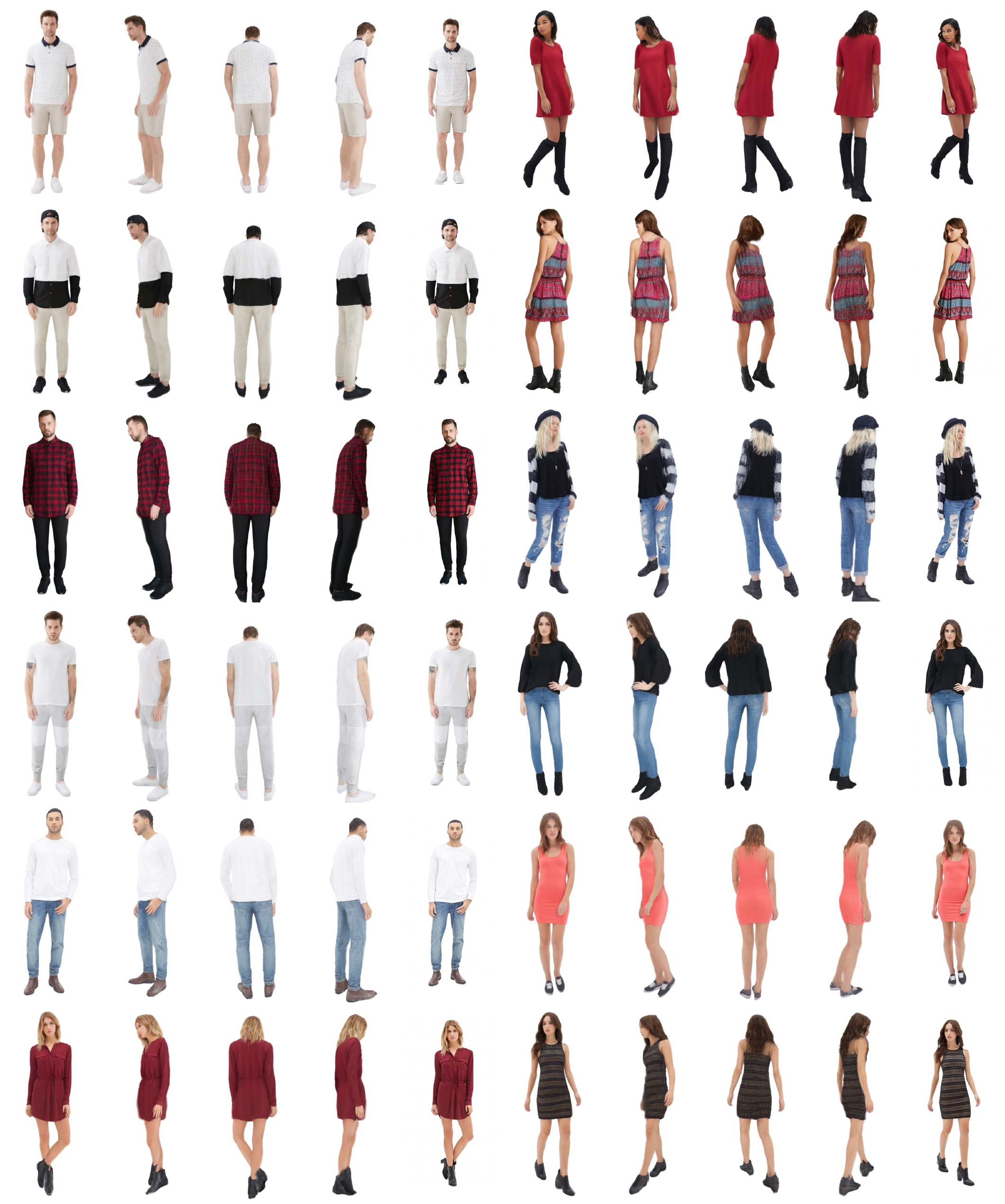}
    \\
    % \makebox[0.01\textwidth]{\footnotesize Multi-View}
    % \makebox[0.2\textwidth]{\footnotesize Reference} 
   % \\ 
    \caption{Visual results of 3D human reconstruction results from a single image~(Part I). Best viewed with zoom-in.} 
    \label{fig:animation_v1}
\end{figure*}

\begin{figure*}[tb] \centering
    \includegraphics[width=0.8\textwidth]{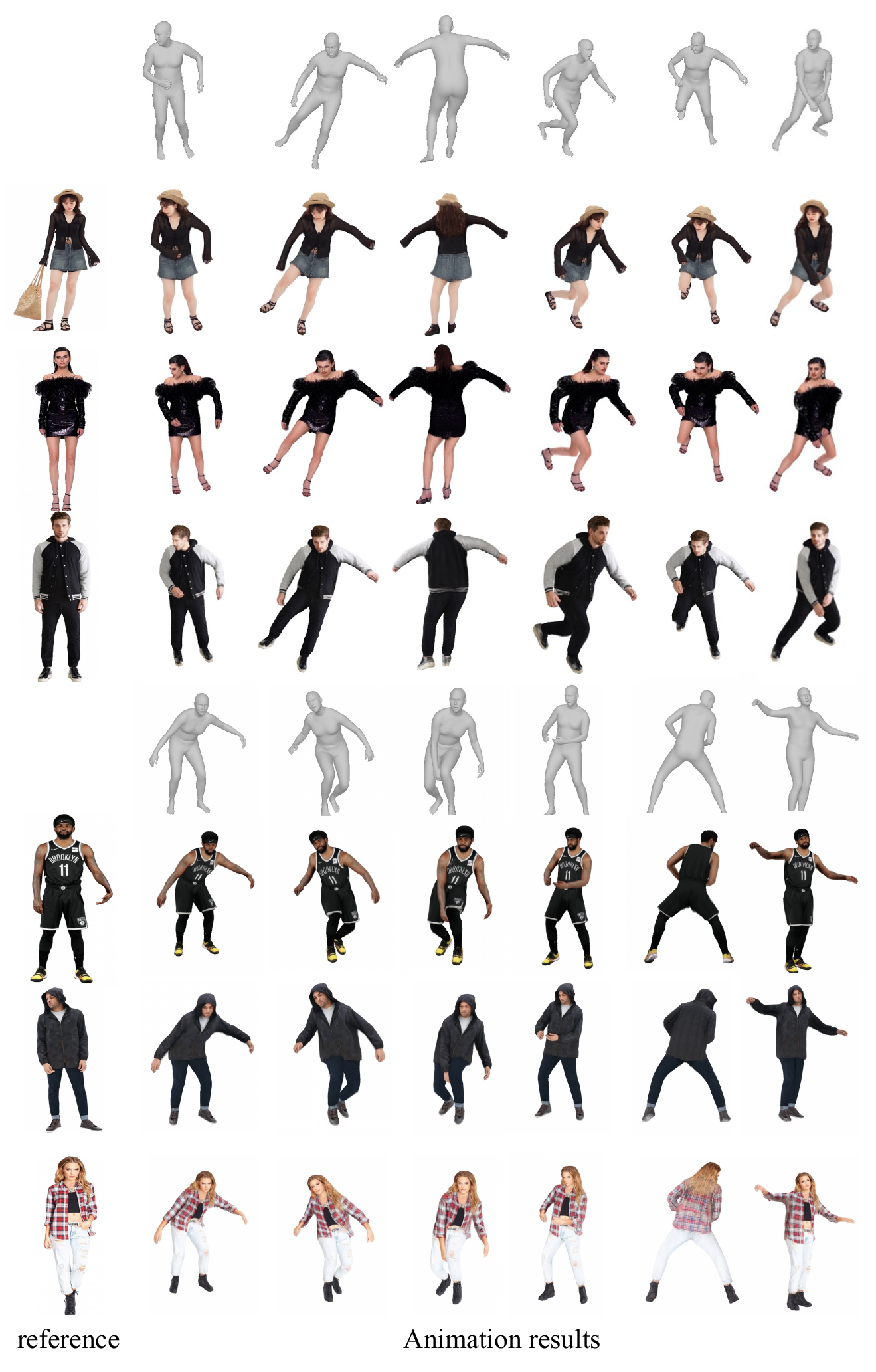}
    \caption{Visual results of 3D human animation from a single image~(Part II). Best viewed with zoom-in.} 
    \label{fig:animation_v2}
\end{figure*}

\Fref{fig:animation_v1}--\Fref{fig:animation_v2} showcase the reconstruction and animation results for input images featuring diverse appearances, clothing, and poses. Our method enables high-fidelity, animatable human avatar reconstruction in a single forward pass with photorealistic rendering, demonstrating its strong generalization and effectiveness.

\end{document}